\documentclass{article}

\PassOptionsToPackage{numbers, compress}{natbib}

\usepackage[preprint]{neurips_2022}

\usepackage[utf8]{inputenc} 
\usepackage[T1]{fontenc}    
\usepackage{hyperref}       
\usepackage{url}            
\usepackage{booktabs}       
\usepackage{amsfonts}       
\usepackage{nicefrac}       
\usepackage{microtype}      
\usepackage{xcolor}         
\usepackage{amsmath}
\usepackage{graphicx}
\usepackage{subcaption}

\usepackage{wrapfig}

\title{Control of Two-way Coupled Fluid Systems with Differentiable Solvers}

\author{%
  Brener Ramos, Felix Trost, Nils Thuerey \\
  Department of Informatics \\
  Technical University of Munich\\
  Boltzmannstraße 3, 85748 Garching bei München, Germany \\
  \texttt{brener.ramos@tum.de, ga94zux@mytum.de, nils.thuerey@tum.de} \\
}

\newcommand{\Reyn}{\operatorname{\mathit{R\kern-.04em e}}}
\newcommand{\baselineTwo}{\textit{BaseNR}\ }
\newcommand{\baselineThree}{\textit{Base}\ }
\newcommand{\buoyancyTwo}{\textit{BuoyNR}\ }
\newcommand{\inflowThree}{\textit{Inflow}\ }
\newcommand{\inflowBuoyancyThree}{\textit{InBuoy}\ }
\newcommand{\hold}{\textit{Hold}\ }

\begin{document}

\maketitle

\begin{abstract}
We investigate the use of deep neural networks to control complex  nonlinear dynamical systems, specifically the movement of a rigid body immersed in a fluid. We solve the Navier Stokes equations with two way coupling, which gives rise to nonlinear perturbations that make the control task very challenging. Neural networks are trained in an unsupervised way to act as controllers with desired characteristics through a process of learning from a differentiable simulator. Here we introduce a set of physically interpretable loss terms to let the networks learn robust and stable interactions. We demonstrate that controllers trained in a canonical setting with quiescent initial conditions reliably generalize to varied and challenging environments such as previously unseen inflow conditions and  forcing, although they do not have any fluid information as input. Further, we show that controllers trained with our approach outperform a variety of classical and learned alternatives in terms of evaluation metrics and generalization capabilities.
\end{abstract}

\section{Introduction}

Control of tasks of physical systems are a ubiquitous challenge in science. In particular, fluids create very difficult environments 
which manifest themselves in simulations via the nonlinearities arising from the Navier Stokes (NS) equations. However, advancements in this field are important for society, and impact areas such as energy, transportation and biology (\citet{Barlas2007, Ho2003, Lord2000}). 

Traditionally, open and closed loop control techniques have been investigated (\citet{Scott2004}).
The latter have clear advantages thanks to their conditioning on state measurements. We investigate and analyze a novel way to train closed loop controllers, namely via deep neural networks recurrently trained in a differentiable simulation environment with physics-based losses. This approach is motivated by the classical challenges of closed loop control for Navier-Stokes environments: fluid flows are complex and chaotic. Moreover, the number of degrees of freedom in numerical solvers is often very large, in turn requiring heavily reduced representations (\citet{Sipp2016, Noack2004, Bergmann2008, Proctor2016}). 
Instead, training with a differentiable simulator provides access to the full, unmodified physical environment, and provides reliable and diverse training feedback in the form of gradients.

More specifically, we investigate steering an actuated rigid body immersed in fluid systems with two way coupling, i.e. the rigid body influences the fluid around it and vice-versa. 
We focus on objectives that require the rigid body to reach specific target configurations, i.e. center of mass location and orientation. In this context, the differentiable simulations make it possible to learn controllers without providing ground truth control signals.

An ubiquitous challenge for neural network approaches is generalization to conditions beyond the training distribution (\citet{Goodfellow2016}). We show that although training takes place in a quiescent flow condition, i.e. a fluid initially at rest, the networks trained via differentiable simulators are able to find control strategies that reliably handle more complex setups than those seen at training time. Their control characteristics are dictated by a set of physically interpretable loss terms, making it possible to favor desired aspects of the control, e.g., the amount of overshoot, tracking speed or maximum control effort. Our networks only receive relative directions in the form of displacement errors, rigid body velocities and previous control efforts. Therefore the neural networks act as a low-to-low dimensional mapping that uses easily accessible sensor data, which mimics potential real world applications. 
The performance of our networks is assessed in four different test scenarios with increasing levels of complexity. We show their advantages over a range of baseline algorithms, from linear controllers such as PID and loop shaping (\citet{mcfarlane1990, Kwakernaak2002}), to supervised and reinforcement learning algorithms (\citet{Haarnoja2018}).

To summarize, our work is the first to investigate deep-learning based controllers using only low dimensional and local information for navigating the nonlinear disturbances of two-way coupled fluid systems. In addition, we make the contribution to demonstrate how a physically interpretable loss function in conjunction with a differentiable solver can be used to train a controller in an unsupervised manner. The resulting controllers not only outperform existing baselines, but also generalize exceptionally well to distinct and new test environments.





\section{Related Work}

Many recent works have been investigating different ways of coupling control and deep learning. Since neural networks are good universal approximators (\citet{HORNIK1989359}), many have investigated using them as a reduced order model of a complex dynamical system (\citet{Eivazi2020, Hasegawa2020, Nair2020}), which can then be used as an inexpensive solver for known closed loop control techniques such as model predictive control (\citet{Bieker2020, Morton2018, Chen2021}). Achieving linear-to-nonlinear mappings through learned Koopman operators has also been studied in recent works (\citet{Yeung2019, Li2020}).

Another way of using deep learning for control purposes is through reinforcement learning (\citet{Verma2018, Paris2021, Feng2021, novati2019, ma2018}). In this case a neural network typically receives a representation of the multi dimensional state describing the system at a given time, e.g. velocity probes and scalar variables, and outputs the control efforts. This is achieved by training the network to maximize a reward function that describes a control objective. In recent years, a variety of refined reinforcement learning variants were proposed (\citet{schulman2015, ho2016generative, Schulman2017,  haarnoja2018soft}). These kind of algorithms traditionally require large amounts of data and training times, which is undesirable especially when considering computationally demanding simulation environments such as fluid simulations.

Recently, differentiable solvers were employed in numerous fields, such as robotics (\citet{Toussaint2019}) and  biology (\citet{ingraham2018}), were constructed to take  advantage of deep learning tools via automatic differentiation.  Since the gradients regarding a cost function are available, it is possible to directly solve for approporiate control efforts of a given task. This task could be placing a piece of cloth into a target container (\citet{Liang2019}), 
pouring liquids (\citet{Schenck2018}), moving a fluid to a specified region (\citet{Holl2020}) or 
generating a specified velocity field from an immersed body  (\citet{Takahashi2021}).
To accomplish the control task, a full optimization needs to be performed for every timestep of a simulation to compute a suitable control signal. 
However, this is typically much too slow for practical applications with real time requirements.
In this work we only use the gradients from the differentiable solver to train a network to act as a controller, which relies only on a sparse set of measurements from the environment. The resulting trained controller can then be evaluated very efficiently. Recent works also investigated using differentiable simulators to accelerate policy learning of various tasks, although generalization capabilities or robustness against disturbances were not assessed (\citet{xu2022}).

\section{Methodology}
\subsection{Governing Equations}
In physics and engineering the evolution of a physical system $\eta(x,t)$ is often described by a partial differential equation (PDE) as

\begin{align}
    \frac{\partial^{n} \eta}{\partial t^{n}} = 
    \mathcal{F} \left( \eta, \frac{\partial \eta}{\partial x},  
    \dotso,
    \frac{\partial^m \eta}{\partial x^m},
    \frac{\partial \eta}{\partial t}, 
    \dotso, 
    \frac{\partial^{n-1} \eta}{\partial t^{n-1}}, 
    \omega(t, \eta) \right) 
    \label{eq:pde}
\end{align}
where $\mathcal{F}$ models the physical behavior of the system and $\omega(t, \eta)$ represents variables that influence it such as boundary conditions. If the system depends only on time as  $\xi(t)$ then \eqref{eq:pde} reduces to an ordinary differential equation (ODE) as
\begin{align}
    \frac{\partial^{n} \xi}{\partial t^{n}} = 
    \mathcal{G} \left( \xi, 
    \frac{\partial \xi}{\partial t}, 
    \dotso, 
    \frac{\partial^{n-1} \xi}{\partial t^{n-1}}, 
    \omega(t) \right) 
    \label{eq:ode}
\end{align}

Given a generic control policy $\hat{\mathcal{P}}(t \mid \theta)$ parametrized by $\theta$, an external actuation can be inserted into a  system described by \eqref{eq:ode}   according to 
\begin{align} \label{eq:xi}
    \frac{\partial^{n} \xi}{\partial t^{n}} = 
    \mathcal{G} \left( \xi, 
    \frac{\partial \xi}{\partial t}, 
    \dotso, 
    \frac{\partial^{n-1} \xi}{\partial t^{n-1}}, 
    \omega(t) \right) 
    + \hat{\mathcal{P}}(t \mid \theta)
\end{align}

In this work, we target a coupled PDE-ODE system, interacting via boundary conditions $\omega$ and the control policy $\hat{\mathcal{P}}$.
Integrating \eqref{eq:xi} over time yields
a state modified by the policy, $\xi(t,\hat{\mathcal{P}})$, and the control task to reach $\xi_{obj}$ is given by the minimization problem
\begin{align}
    \arg \min_{\theta } \| \xi_{obj} - \xi(t, \hat{\mathcal{P}}(t \mid \theta)) \| .
\end{align}

More specifically, we use the incompressible Navier Stokes equations, which is a form of \eqref{eq:pde} with $n=1$, that describes how a velocity field evolves given specified boundary conditions as the following   
\begin{equation}
    \frac{\partial u}{\partial t} = - u \cdot \nabla u - \frac{\nabla p}{\rho} + \nu \nabla^2u
    \label{eq:ns}
\end{equation}
where $u$ is the velocity field, $p$ is the pressure, $\rho$ is the density and $\nu = \frac{\hat{u} \, \hat{L}}{\Reyn}$ is the kinematic viscosity, where $\Reyn$ is the Reynolds number and $\hat{u}$ and $\hat{L}$ are a reference velocity and length, respectively. A Poisson equation is also solved for the pressure in order to enforce the velocity field to be divergence free. 

We additionally target rigid objects immersed in the fluid.
Their linear and angular movement can be described by \eqref{eq:ode} with $n=2$ as
\begin{align}
    \frac{\partial^2  {x_r}}{\partial t^2} & = \frac{1}{m} \sum F  \label{eq:rbl} \\
    \frac{\partial^2  {\alpha}}{\partial t^2} & = \frac{1}{I} \sum T \label{eq:rba}
\end{align}
where $x_r$ is the body position, $m$ the body mass, $\alpha$ the body angle and $I$ its moment of inertia. 
The terms $\sum F$ and $\sum T$ denote  the forces and torques that are acting upon the body, respectively. When using both \eqref{eq:rbl} and \eqref{eq:rba} the system has 3 degrees of freedom (DOF). In a few cases below we will omit \eqref{eq:rba}, yielding a simplified 2 DOF scenario.
When dealing with a rigid body immersed in a fluid, these terms reduce to 
\begin{align}
    \sum F &= - \oint_S p \, \vec{n}(s) \; ds  \label{eq:linear_acc} \\
    \sum T &= - \oint_S \vec{r}(s) \times p \, \vec{n}(s) \, ds  \label{eq:ang_acc}
\end{align}
where $S$ is the body surface, $\vec{n}$ is the surface normal, $\vec{r}$ maps the surface location to the local coordinate system of the body, with the origin being the center of mass. 
%
Dirichlet boundary conditions for the NS simulation are imposed on the velocity field via
$u_{\Omega} = \partial x_r/ \partial t + \partial \alpha/ \partial t \; r_{\Omega}$ at rigid body surface cells defined by contour $\Omega$.

Together, \eqref{eq:rbl} and \eqref{eq:rba} represent a coupled dynamical system subjected to nonlinear perturbations derived from the interaction between rigid body and fluid. 
Control efforts after exerted via forces:
\begin{align}
    \sum F &= - \oint_S p \, \vec{n}(s) \; ds  + F_c   \\
    \sum T &= - \oint_S \vec{r}(s) \times p \, \vec{n}(s) \, ds  + T_c
\end{align}

The specific control problem can then be formulated by finding the control efforts through $[F_c, \, T_c]^T = \mathcal{P}(t \mid \theta)$ so that
\begin{align}
\arg \min_{\theta}  \|e_{xy} \|  + \|e_{\alpha}   \| \label{eq:mini} \\ 
    e_{xy} =  x_{obj} - x_r(t, \mathcal{P}(t \mid \theta)) \\ 
    e_{\alpha} =  \alpha_{obj} - \alpha (t, \mathcal{P}(t \mid \theta))
\end{align}
where $x_{obj}$ and $\alpha_{obj}$ are an objective position and angle, respectively. Therefore the control task investigated can be summarized as controlling an ODE (rigid body movement equations) with highly nonlinear disturbances that emerge from a PDE (NS equations), which is also influenced by the ODE solution. 

\begin{wrapfigure}{hR}{0.4\linewidth}
    \vspace{-2mm} 
    \centering
    \includegraphics[width=0.99\linewidth]{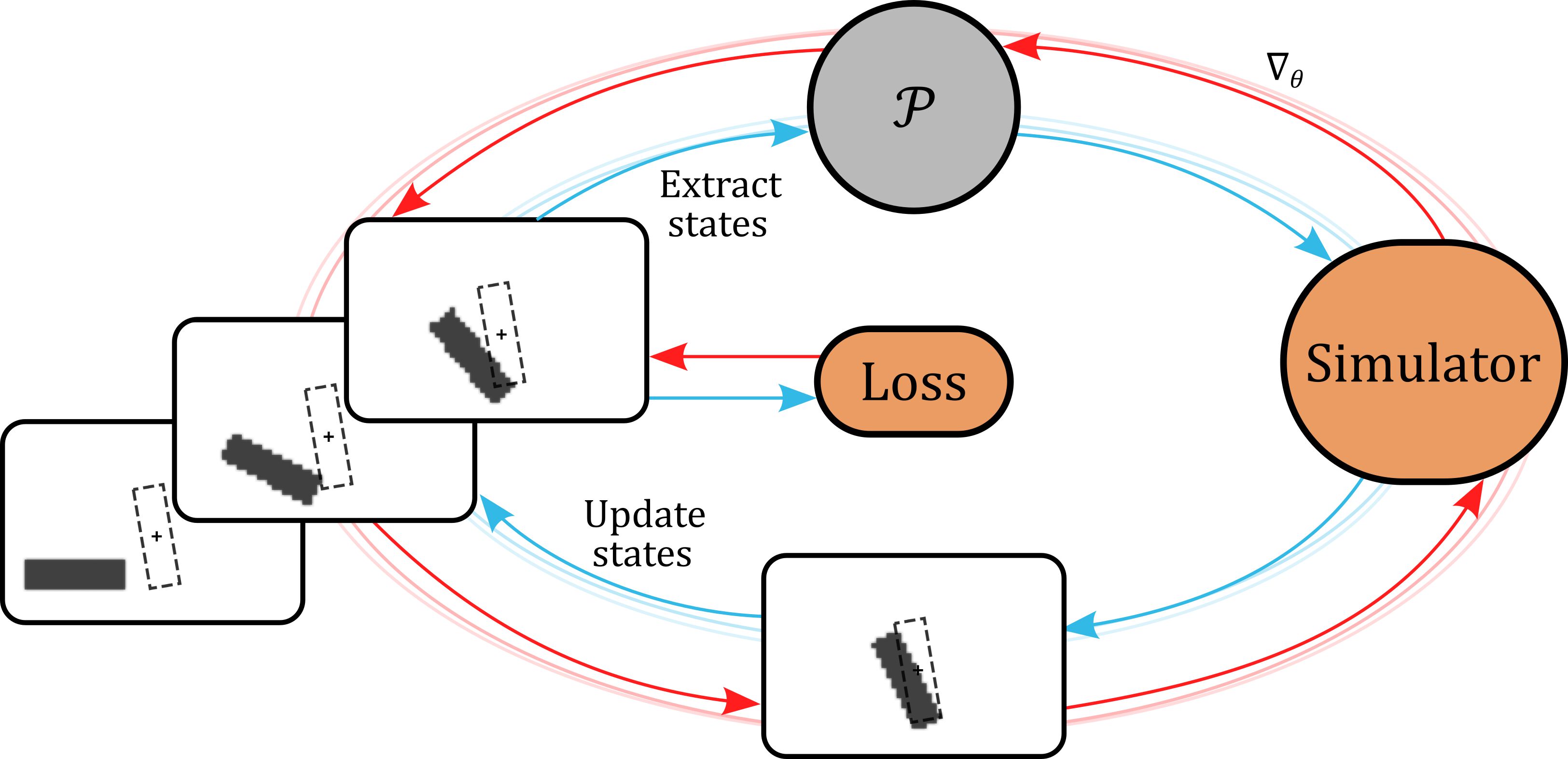}
    \caption{\footnotesize{
    A schematic of the differentiable solver training:
    Blue arrows represent the forward pass while red ones illustrate the flow of gradients.
    Importantly, the loss signal is backpropagated through $l$ simulation steps to provide policy $\mathcal{P}$ with long-term feedback about the flow environment.
    }
    } \label{fig:network}
    \vspace{-2mm} 
\end{wrapfigure}

\subsection{Differentiable Solver}
Our differentiable fluid solver
is based on Chorin-projections with a second-order advection operator. Signed distance functions of the moving obstacles are rasterized to the Eulerian simulation grid to flag cells as being either a fluid or an obstacle. Dirichlet BCs as well as isolating the rigid body surface are set in accordance to this cell classification in order to maintain differentiability. 
We implement our solver via the PhiFlow framework (\url{https://github.com/tum-pbs/PhiFlow}) using PyTorch as backend.

Each time step of the solver requires solving a Poisson’s problem for the pressure considering the rigid body as boundary conditions. In order to avoid noisy gradients during backpropagation, the solver uses a custom PyTorch autograd function that solves a linear system in both backward and forward pass. The solid-fluid coupling is realized via a two-way coupling where NS and rigid body equations are solved in an alternating fashion. The body influences the fluid by changing the fluid’s BCs, and the fluid acts on the rigid body by exerting force and torque calculated with Equations \ref{eq:linear_acc} and \ref{eq:ang_acc}. In this way, the control efforts gathered from the neural networks are propagated through the simulation graph, allowing gradients to flow from the loss function to the network weights.

\section{Neural Networks as Control Policies}


We investigate how to use neural networks to represent a policy $\mathcal{P}(z(t) \mid \theta)$ for the control task described by \eqref{eq:mini}, where $z(t)$ is a set of discrete low dimensional state variables,
which we denote with $z$ for brevity.
The network with weights $\theta$ acts as a policy, which receives the
current and previous $n_p$ states as input. It has the task to infer appropriate control efforts for a given learning objective.
Each state is a low-dimensional array that consists of the spatial error $e_{xy}$, angular error $e_{\alpha}$, rigid body linear velocity $\frac{\partial x_r}{\partial t}$, rigid body angular velocity $\frac{\partial \alpha}{\partial t}$ and control forces $F_c$ and torques $T_c$. 
Also $\frac{\partial x_r}{\partial t}$ and $\frac{\partial \alpha}{\partial t}$ will be referred to as  $\Dot{x}$ and $\Dot{\alpha}$, respectively. 
$e_{xy}$, $\Dot{x}$ and $F_c$ are expressed in the local reference frame of the rigid body. 
Therefore no global information describing the state of the fluid is transmitted to the control policy.
We denote states at a discrete time t with a superscript $^t$.
In this way, the input for the networks at a time $t$ can be expressed as $z =
[w^t, \, w^{t-1}, \, \dotso, w^{t-n_p} ] $, where $w^t = [ e_{xy}^t, \, \Dot{x}^t, \, F_c^t, \, e_{\alpha}^t, \, \Dot{\alpha}^t, \, T_c^t]^T $. In the following, a range of different learning procedures are investigated: training with differentiable physics simulators, a fully supervised approach, and a reinforcement learning variant.

    
%

\subsection{Learning via a Differentiable Solver}
\label{sec:ol}

Our method employs a fully differentiable solver which allows gradients to be provided to the neural network policy $\mathcal{P}_{\text{diff}}(z \mid \theta)$ about reactions of the physical system regarding previous policy actions and its temporal evolution. 
This policy can be trained without the need to pre-compute potentially sub-optimal training data. 
Rather, the network is left to discover the best possible policy over the course of the training in an unsupervised way.
Our loss formulation includes a time horizon of $l$ time steps as a central parameter (details on how we choose its value can be found in Section \ref{sec:time_horizon}). The evaluations across this time interval leads to training signals that take into account how outputs of the policy network influence the future states of the environment. Via the differentiable solver, the loss signals are recurrently backpropagated to the policy, making it more ``aware'' of the consequences of previous actions.
This process is illustrated in Figure \ref{fig:network}.
We make use of a loss function that combines three objectives. The objective term, $O$, typically dominates, and ensures that the body reaches the target state:
\begin{align}
    O = \frac{\beta_{xy}}{l}\sum_{n=0}^{l-1}\|e_{xy}^n\|^2 +
        \frac{\beta_{\alpha}}{l}\sum_{n=0}^{l-1} \|e_{\alpha}^n\|^2 , 
\end{align}
where the $\beta$ are hyperparameters that weigh the different terms. However, a loss function with this term alone results in a controller with tendencies of overshoot since it does not account for the rigid body velocity. Hence we introduce a velocity term $V$:
\begin{align}
    V = \frac{\beta_{\Dot{x}}}{l}\sum_{n=0}^{l-1}\frac{\|\Dot{x}^n\|^2}{\beta_{prox}\|e_{xy}^n\|^2 + 1} +  
    \frac{\beta_{\Dot{\alpha}}}{l}\sum_{n=0}^{l-1}\frac{\|\Dot{\alpha}^n\|^2}{\beta_{prox}\|e_{\alpha}^n\|^2 + 1}   
\end{align}
Far away from the target, larger spatial and angular errors in the denominators lead to smaller values of $V$.
Closer to the target objective these errors approach zero, and hence $V$ becomes an L2 norm of the linear and angular velocities. As a consequence the optimization guides the policy to slow down the body only when near the target, thus reducing overshooting effects. 
%
Following previous work \citep{Bieker2020}, we additionally include a term to avoid large control efforts as well as abrupt changes
\begin{align}
    E & = \frac{\beta_F}{l}\sum_{n=0}^{l-1} \|F_c^n\|^2
    + \frac{\beta_T}{l}\sum_{n=0}^{l-1} \|T_c^n\|^2 +  \notag\\ 
    &  \frac{\beta_{\Delta F}}{l}\sum_{n=0}^{l-1} \|F_c^n-F_c^{n-1}\|^2
    + \frac{\beta_{\Delta T}}{l}\sum_{n=0}^{l-1} \|T_c^n-T_c^{n-1}\|^2 
\end{align}

Finally the combined loss function for the differentiable solver training can be written as 
\begin{align}
    L = O + V + E \label{eq:loss}
\end{align}
Above, the $\beta$ hyperparameters regulate the relative impact of each term for the controller. 
Because of the direct physical impact of each hyperparameter, adjusting them is straight forward.
For example, spatial tracking can be made more precise by increasing $\beta_{xy}$ or the angular tracking can be accelerated by decreasing $\beta_{\Dot{\alpha}}$. 
However, overly large values can result in overshooting. An ablation study can be found in Section \ref{sec:ablation}.
The hyperparameters for this work were chosen aiming for an overall balance of the objectives 
. 

\subsection{Supervised Learning}

As a baseline for learning, we include a fully supervised learning approach.
Due to a lack of optimal, ground-truth control policies, we construct a dataset in the following manner:
we manually prescribe velocities to the rigid body so that it reaches an arbitrary target. We then compute the fluid forces acting upon the body and calculate the control efforts to cancel them and yield  the acceleration of the prescribed trajectory. In this way, we obtain a set of states $z$ with paired, expected control efforts $[\hat{F}_c(z), \, \hat{T}_c(z)]^T$.
Training with these precomputed forces can be performed in a fully supervised way with the loss 
\begin{align} \label{eq:supervised}
  L = \| [\hat{F}_c(z), \, \hat{T}_c(z)]^T - \mathcal{P}_{\text{sup}}(z \mid \theta ) \|^2 .  
\end{align}
%

\subsection{Reinforcement Learning}
Additionally, we include a reinforcement learning algorithm that works without making use of the solver gradients. We use the Soft Actor Critic (SAC) algorithm \citep{haarnoja2018soft}, a recent state-of-the-art approach. SAC is a model-free variant that has the added benefit of being off-policy. Thus, past experiences can be stored in a replay buffer and are not invalidated by policy updates. As the simulation process is computationally very expensive, this increase in sample efficiency is highly beneficial.
The actor, which is also represented by a neural network, has the objective of maximizing the predicted Q values. Additionally, SAC introduces an entropy term into the objective of the policy as regularization. It discourages overly confident decisions while also controlling the trade-off between exploration and exploitation inside the action space. 
In our setting, the actor represents a control policy, and hence we refer to it as $\mathcal{P}_{\text{RL}}(z \mid \theta)$. 
The reward function uses the same formulation as the differentiable physics case, and is computed by multiplying equation \ref{eq:loss} by -1.

\section{Experiments}

We perform a series of experiments with increasing degrees of complexity to assess the generalization and stability capabilities and stability of the considered approaches. Below we explain the default parameters which are applicable unless noted otherwise. Details and deviating parameters for all experiments are provided in the appendix. Source code used in this work will be provided upon acceptance.

\begin{wraptable}{r}{8.5cm}
\caption{Main parameters of experimental setups.}
\vspace{-0.3cm}
\label{tab:tests}
\begin{center}
\begin{small}
\begin{tabular}{cccccccc}
\toprule
 \textbf{ID}  &  \textbf{Inflow} & \textbf{Buoy.} & \textbf{Forcing}  & \textbf{DOF} & $\Reyn$
\\ \hline \\
\baselineTwo  &     $-$ & $-$ & $-$ & 2 & 1000\\
\buoyancyTwo  &     $-$  & $\surd$ & $-$ & 2 & 1000\\
\baselineThree  &    $-$ & $-$ & $-$ & 3 & 1000\\
\inflowThree  &     $\surd$ & $-$ & $-$ & 3 & 3000\\
\inflowBuoyancyThree  &  $\surd$  & $\surd$ & $-$ & 3 & 3000\\
\hold  &    $\surd$  & $\surd$ & $\surd$ & 3 & 3000\\
\bottomrule
\end{tabular}
\end{small}
\end{center}

\end{wraptable}

\paragraph{Datasets and Test Scenarios}

The training data consists of simulations with only one objective in the form of a target configuration,
and uses the standard sets of parameters provided in Table~\ref{tab:tests}.
The two baseline versions are denoted by \baselineTwo, indicating "no rotation",for systems with 2 DOF, and \baselineThree for 3 DOF. 
All training samples use a quiescent flow ($u=0$) as initial condition. 
Since the fluid is initially at rest, all perturbations are created from the rigid body movement. 
Validation datasets  consist of 20 simulations with the same parameters as training, i.e. \baselineTwo and \baselineThree, but different objectives. 
To evaluate generalization of 2 DOF networks, simulations with parameters \buoyancyTwo are performed, which increase the control task difficulty by introducing a lighter fluid source that disrupts the flow through buoyancy.
For 3 DOF networks trained on \baselineThree, we use the environments \inflowThree, \inflowBuoyancyThree and \hold. For \inflowThree, an inflow is present, the fluid is less viscous (higher $\Reyn$) and $\Delta t$ is smaller. Correspondingly, the controllers are sampled once every two timesteps. 
The environment \inflowBuoyancyThree adds a lighter fluid source at the bottom of the domain that disrupts the flow through buoyancy. Finally, simulations with \hold parameters have additional  prescribed forcing at certain moments. 
These test environments were designed to deviate more and more strongly from the quiescent training conditions, making the control task progressively harder. 
These changes in environments are not directly transmitted to the controllers, rather, they should adapt to the new conditions based on their training.
Unless stated otherwise, the test datasets are comprised of 5 simulations with different trajectories created by changing the objectives after $\Delta t=100$ when considering \buoyancyTwo, \inflowThree and \inflowBuoyancyThree. Tests in the \hold environment are comprised of a single simulation
which, in contrast to the other cases, has the goal to keep and stabilize the rigid body in the initial position. This is made more difficult by exerting additional forcing on the body.
The supervised learning approach uses a separate, pre-computed dataset for training, which is comprised of 100 simulations from which 80 are used for training and 20 for validation and each one has 500 time steps.



\paragraph{Neural Network Representation and Training}
The underlying network architecture for all approaches is kept constant: two dense layers with ReLU activation, followed by a third dense layer. For the latter, the networks trained via differentiable physics and reinforcement learning have a hyperbolic tangent activation to ensure that the control efforts are bounded. 
As we achieved a better performance for the supervised case without the activation of the last layer, it was omitted there. 
Training the differentiable physics networks starts with a simulation with quiescent initial condition, gathering objectives from an uniform random distribution. After advancing the simulation so that $n_p$ past states exist, the network training is activated and its outputs are used as control efforts. Once $l=16$ time steps are available, we compute a loss and update $\theta$. 
A discussion about how we choose $l$ can be found in \ref{sec:ablation}. 
After 1000 simulation steps we restart the simulation with new targets until $n_i$ training iterations are performed.
We choose $n_i=1000$ and $n_i=5000$ when training a network for 2 and 3 DOF systems, respectively. 
%
%
The reinforcement learning approach uses the same simulation environment as the differentiable training but without making use of variables' gradients. After each simulated time step a batch of 128 samples is drawn from the replay buffer and used for training. 
In total, $n_{i}=500{,}000$ iterations are performed and the validation error of intermediate models are computed. Furthermore, the model with the lowest validation error was chosen for further testing.



\paragraph{Evaluation Metrics} 
As error metrics for comparing results we primarily use absolute errors in position and orientation, $||e_{xy}||$ and $||e_{\alpha}||$.
We also compute an average steady state error $\overline{\| e \|}_{ss}$ that can assess the steady-state performance of the controllers without their initial transient phases. 
It is calculated using the last $3/4$ of the time interval in which an objective was being tracked. However, since the rigid body starts in a null error position when considering tests in the \hold environment, the steady state error for these cases is the whole time average of position and orientation errors.
For bar plots of $\overline{\| e \|}_{ss}$ we also display the standard deviation for illustrating the spread of the error values.


\section{Results}


\subsection{Algorithmic Comparison}
%

\paragraph{Baseline Algorithms} 
Two types of linear controllers are tested as a reference to help assess the performance of the learned versions: a classic proportional-integrator-derivative (PID) $\mathcal{P}_{\text{PID}}$ and a loop shaping controller $\mathcal{P}_{\text{LS}}$ designed through a blend of mixed-sensitivity design and the Glover-McFarlane method (\citet{mcfarlane1990, Kwakernaak2002}). 

\begin{wrapfigure}{hR}{0.5\linewidth}
    \centering
    \begin{subfigure}{\linewidth}
        \centerline{\includegraphics[width=\linewidth]{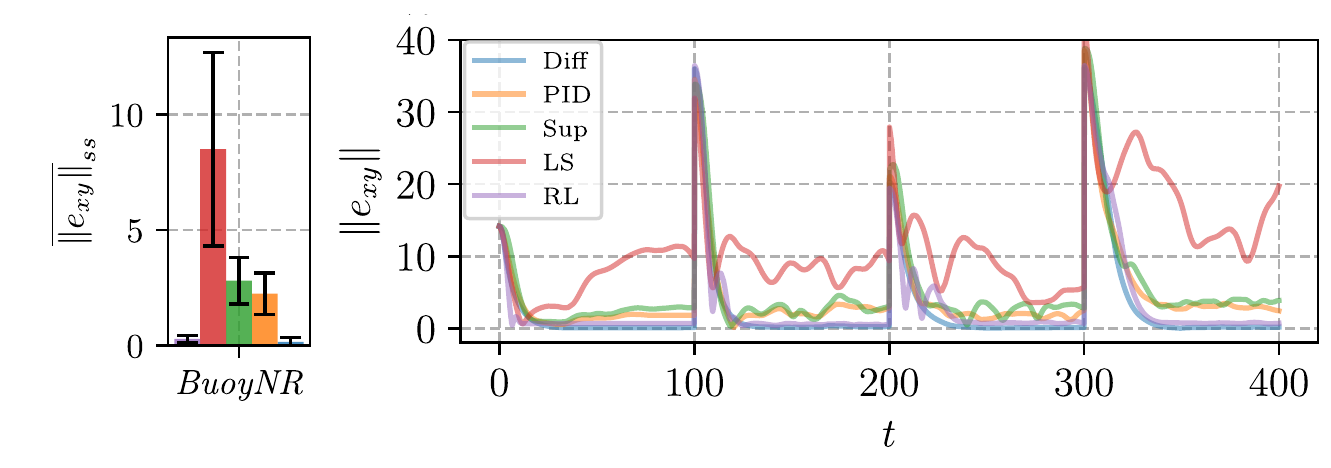}}
    \end{subfigure}
    
    \begin{subfigure}{\linewidth}
        \centerline{\includegraphics[width=\linewidth]{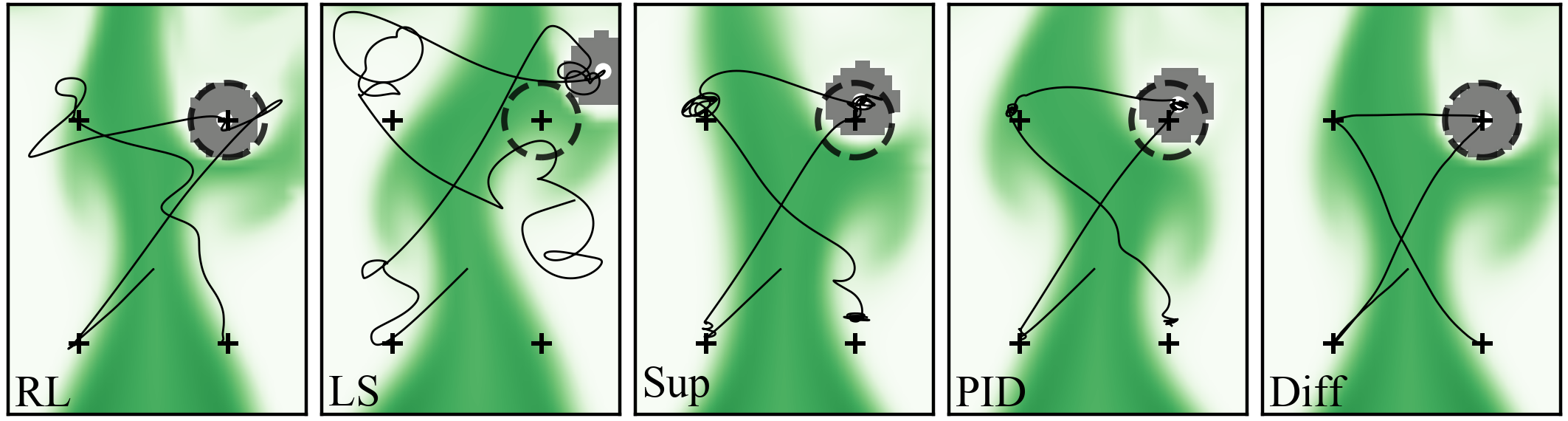}}
    \end{subfigure}
    \caption{Average spatial steady state errors top-left), trajectories of one of the test simulations (botom) and their error norms (top-right) with parameters \buoyancyTwo. When compared to the other approaches $\mathcal{P}_{\text{diff}}$ exhibit less oscillations and a lower average steady state error.}
    \label{fig:buoy2}
\end{wrapfigure}

\paragraph{2 DOF Validation} 
First, we consider a validation with the setup \baselineTwo (same simulation parameters as training but with different targets). 
The goal is to make sure all the controllers are functioning correctly for the conditions seen at training time. All the approaches considered are able to achieve a low error given sufficient time, which confirms they work as intended in conditions they were designed to operate in. Details can be found in the appendix (Figure \ref{fig:ba2}).

    

\paragraph{Increased Difficulty}
Next we perform a test with the setup \buoyancyTwo, which includes perturbations from a lighter fluid, different $\Reyn$ 
and a longer time window. 
The interaction between buoyancy and rigid body creates oscillatory flow structures and the control task becomes much more challenging. As a consequence, $\overline{\| e_{xy} \|}_{ss}$ is much larger for all approaches except for  $\mathcal{P}_{\text{RL}}$ and $\mathcal{P}_{\text{diff}}$. 
The latter however has a more stable trajectory, as shown in Figure \ref{fig:buoy2}.
A large amount of undesirable oscillations and overshooting is present when using all other controllers.
It is worth noticing that  $\mathcal{P}_{\text{RL}}$ requires very long training times when compared to the other learned approaches. E.g., training  $\mathcal{P}_{\text{RL}}$ takes 35 times longer to train than $\mathcal{P}_{\text{diff}}$.
Additionally, $\mathcal{P}_{\text{RL}}$ training exhibits a high variance in performance, making it necessary to assess the performance of many intermediate models in order to find a stable and suitable one. For $\mathcal{P}_{\text{LS}}$, it can be seen that it produces the largest errors. For these reasons we will omit  $\mathcal{P}_{\text{RL}}$ and $\mathcal{P}_{\text{LS}}$ for further tests.

\subsection{Increased Complexity and Generalization}

We now evaluate the performance of the remaining methods for a rectangular body with rotation (3 DOF).
The additional degree of freedom adds a significant amount of complexity to the control task, and
we use a set of more complex test cases to evaluate generalization.
%
%
In line with the 2 DOF case, we first perform an evaluation with validation cases with the same simulation parameters as training (\baselineThree). All three remaining methods fare well for these environments, and yield stable controllers. Details are provided in the appendix (Figure \ref{fig:base3}).
%

\begin{wrapfigure}{hR}{0.6\linewidth}
    \centering
    %
    %
    \begin{subfigure}{0.18\linewidth}
        \centerline{\includegraphics[width=\linewidth]{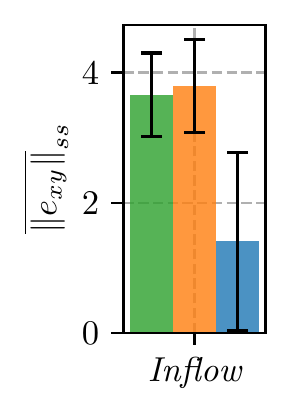}}
    \end{subfigure}
    \begin{subfigure}{0.81\linewidth}
        \centerline{\includegraphics[width=\linewidth]{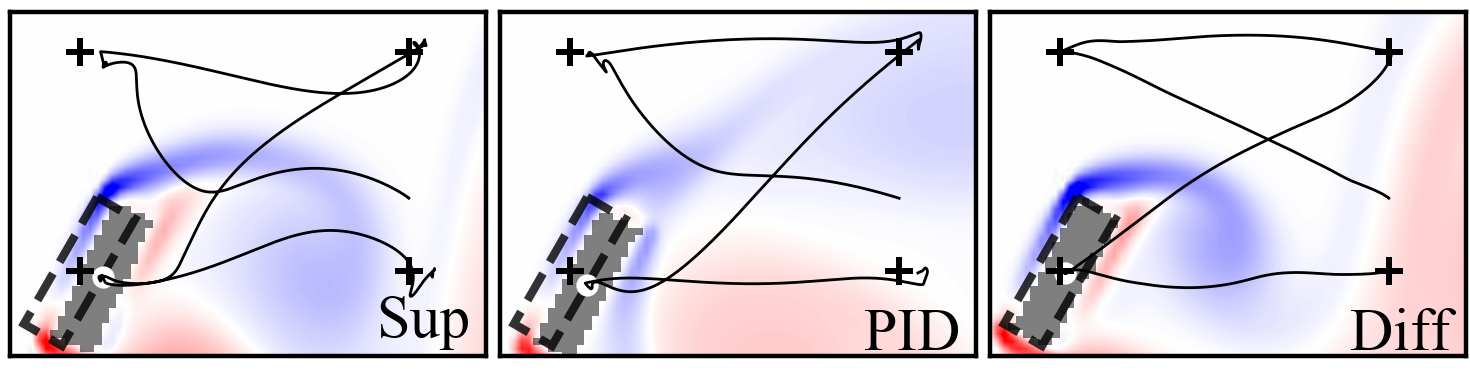}}
    \end{subfigure}

    \caption{Average steady state errors (left), trajectories of one of the test simulations (right) with parameters \inflowThree. Trajectories are colored by vorticity ($t=295$). $\mathcal{P}_{\text{sup}}$ and $\mathcal{P}_{\text{PID}}$ exhibit an offset when trying to reach target positions as well as mild oscillatory trajectories. Those behaviors are not observed when deploying $\mathcal{P}_{\text{diff}}$.}
    \label{fig:inflow3}
\end{wrapfigure}

\paragraph{Generalization Tests} 
First, tests with the \inflowThree setup are performed to assess the generalization capabilities of the considered controllers. 
Here the environments are changed substantially by introducing an inflow from the left side of the domain and utilizing a smaller viscosity (higher $\Reyn$). The inflow together with the rigid body movement creates unsteady flow structures that vary depending on the box angle.  $\mathcal{P}_{\text{diff}}$ is able to maintain a smaller $\overline{\| e_{xy} \|}_{ss}$ compared to the other controllers as can be seen in Figure \ref{fig:inflow3}. 
Very noticeable oscillations around the objective locations are present in the trajectory of $\mathcal{P}_{\text{sup}}$ and $\mathcal{P}_{\text{PID}}$. Instead, $\mathcal{P}_{\text{diff}}$ produces a stable orientation, showing that this policy successfully counteracts the perturbations caused by the strongly varying flow.

\begin{figure}[!h]
    \centering
    \begin{subfigure}{0.49\linewidth}
        \centerline{\includegraphics[width=\linewidth]{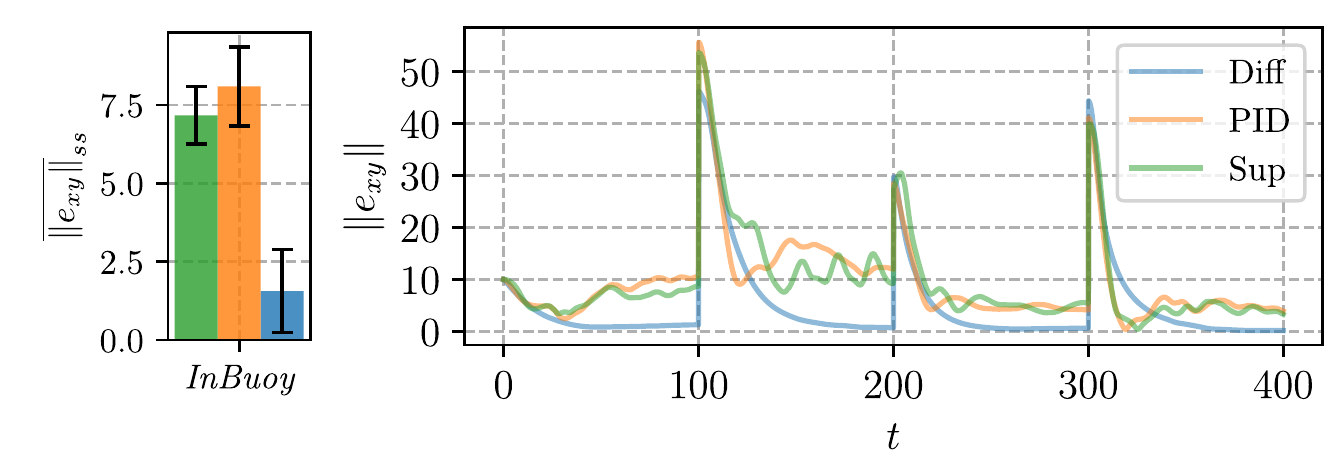}}
    \end{subfigure}
    %
    %
    \begin{subfigure}{0.49\linewidth}
        \centerline{\includegraphics[width=\linewidth]{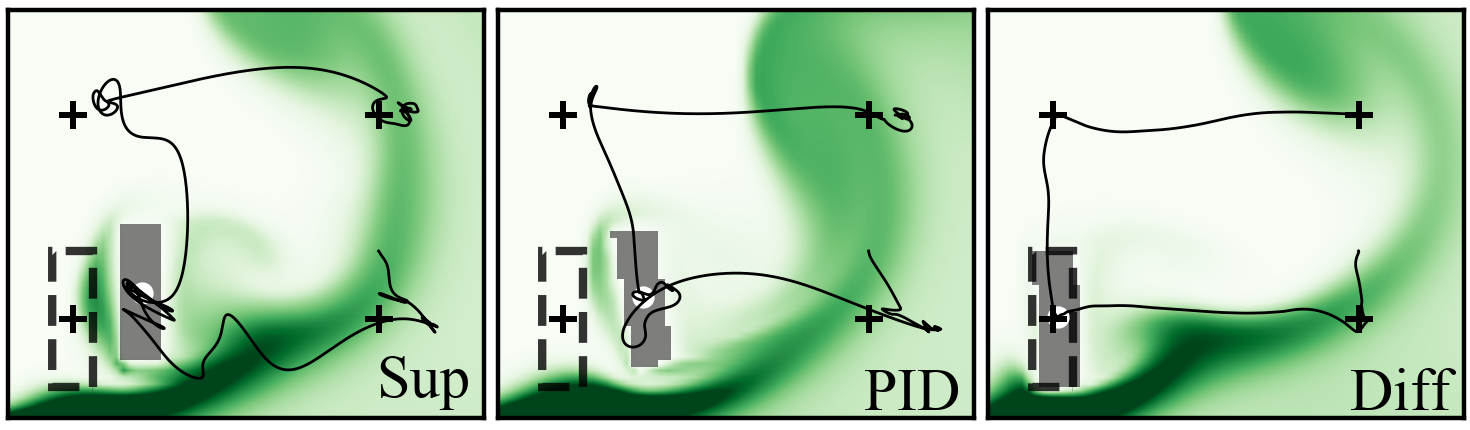}}
    \end{subfigure}
    \vskip -0.1in
    \caption{Average steady state errors (left), trajectories of one of the test simulations (right) and their error norms (middle)  with parameters \inflowBuoyancyThree. Trajectories are colored by lighter fluid ($t=195$). Buoyancy worsens the performance of controllers but $\mathcal{P}_{\text{diff}}$ still maintains low steady state errors.}
    \label{fig:inflowbuoyancy3}
\end{figure}

%
\textit{Inflow and Buoyancy:} 
To increase difficulty, we introduce a source of lighter fluid that rises due to buoyancy at the bottom of the domain with parameters \inflowBuoyancyThree.
This test has the goal of assessing the robustness of the controllers further, since the fluid source tends to create higher frequency oscillations, which makes the control task harder.  $\mathcal{P}_{\text{sup}}$ and $\mathcal{P}_{\text{PID}}$ show a considerable worsening of their performance with higher steady state errors and undesirable oscillations. On the other hand, $\mathcal{P}_{\text{diff}}$ maintains low values for $\overline{\| e_{xy} \|}_{ss}$  and $\overline{\| e_{\alpha} \|}_{ss}$  while successfully suppressing most of the perturbations as shown in Figure \ref{fig:inflowbuoyancy3}. 

\begin{figure}[!h]
    \centering
    \begin{subfigure}{0.49\linewidth}
        \centerline{\includegraphics[width=\linewidth]{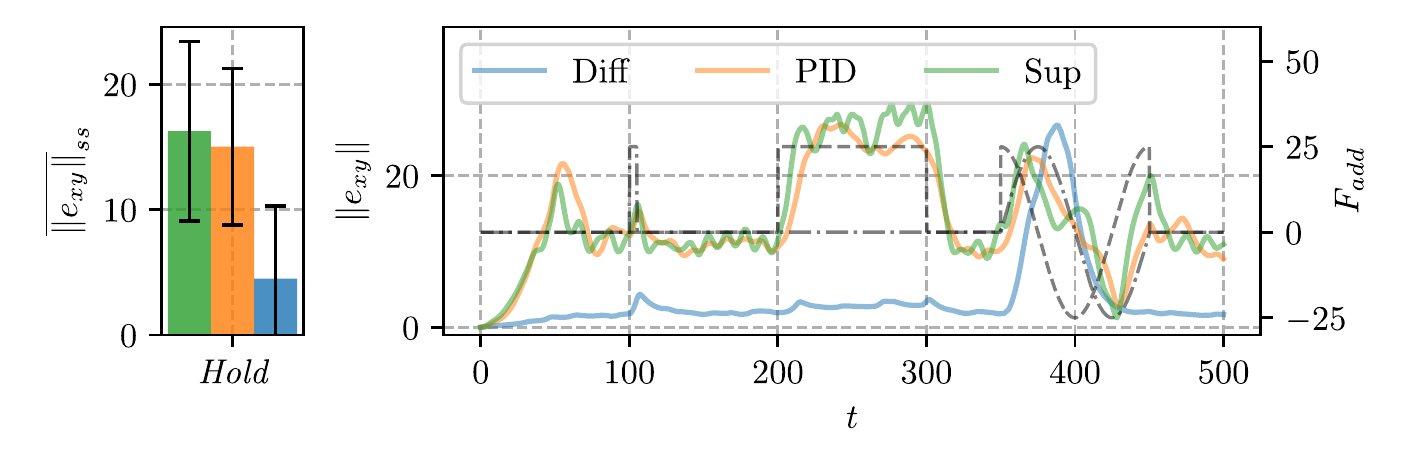}}
    \end{subfigure}
    %
    %
    \begin{subfigure}{0.49\linewidth}
        \centerline{\includegraphics[width=\linewidth]{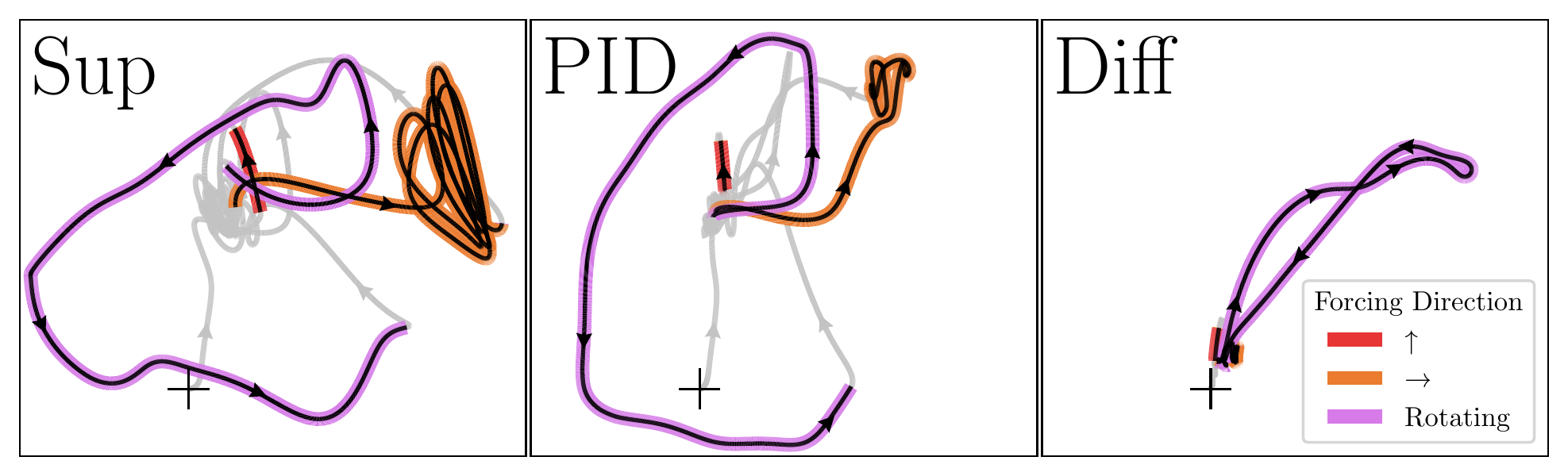}}
    \end{subfigure}
    \caption{Average steady state errors (left), center-of-mass trajectories of simulations colored by forcing type (right) and their error norms (middle) for the \hold environment. Additional forcing in $x$ and $y$ axes are represented by dashed and dotted-dashed black lines, respectively. Controllers must maintain the rigid body at the target location despite perturbations. As visualized on the right, $\mathcal{P}_{\text{diff}}$ is able handle vertical and horizontal  forcing (red and orange) more efficiently than $\mathcal{P}_{\text{sup}}$ and $\mathcal{P}_{\text{PID}}$. Tracking briefly worsens when rotational forcing is applied (purple), due to maximum control efforts being smaller than the fluid forces combined with forcing.}
    \label{fig:inbuofor}
\end{figure}

%
\textit{Hold:}  
The last test introduces additional forcing to the rigid body in addition to the disturbances from the fluid. Since the simulation starts with the rigid body at the target position and orientation, the controllers primarily need to counter the external forces. $\mathcal{P}_{\text{diff}}$ is able to 
counteract the additional forcing,  
with the exception of a brief lapse at $t=390$ as shown in Figure \ref{fig:inbuofor}. This is caused by the fact that the fluid forces combined with additional forcing are momentarily larger than the maximum control forces. However $\mathcal{P}_{\text{diff}}$ is able to recover and lock into the target position again afterwards. Although $\mathcal{P}_{\text{sup}}$ and $\mathcal{P}_{\text{PID}}$ have unbounded maximum control efforts, they are not able to stabilize the rigid body at the target position.


\subsection{Discussion}

Taken together, the previous set of tests show the advantages of the proposed training via differentiable simulations.
It is able to provide a neural network with feedback to control a dynamical system subjected to nonlinear perturbations for long periods of time, even though training is performed in a different, simplified environment. By backpropagating through $l$ steps, the gradient of the time evolution of the system greatly helps the optimization process to find useful relations between the input measurements in order to generate appropriate controls while not relying on inputs containing global information.
This is especially important for finding non-trivial control strategies, or if a reference dataset of control signals can not be provided. Our results show that complex control strategies can be learned in an unsupervised manner as long as the control objective and controller characteristics can be expressed in a mathematically meaningful way.

Another interesting characteristic is the robustness against the disturbances seen in the tests. 
Although some tracking performance is lost the more complex the tests become, 
$\mathcal{P}_{\text{diff}}$ is able to handle new situations without the strongly deteriorated tracking performance of the other controllers.
Our tests indicate that this is a consequence of the differentiable solver, which provides a varied learning signal and results in a neural network that robustly handles a large variety of conditions.

It is also conceptually very appealing for control algorithms to be able to use the original set of 
equations of a system, even if they are as nonlinear as the NS equations. This is due to the fact that
reduced representations often do not portrait  nonlinear and unexpected behavior well enough. Instead, the differentiable simulation approach allows training the controller efficiently using the full set of model equations, such that it can adapt to the subtleties of the environment.

On the other hand, a drawback of the proposed approach is the fact that differentiable simulations lead can lead to increased training times because the simulation has to be evaluated in the forward, as well as the backward pass.
In our experiments we found this can be up to 4 times slower than training via a supervised policy. The latter, however, induces an additional cost for generating the training data, which is likely to be substantial. 
The differentiable solver training shares the increased training cost with reinforcement learning, but a single iteration of the former is roughly twice as expensive due to the backwards path for calculating the gradient of the simulator. However, the gradients typically lead to faster convergence. In our tests, the reinforcement learning approach on average takes an order of magnitude longer to train and exhibits high variance in terms of models performance.
Yet, the resulting neural network is equally fast to evaluate for all three cases, since it relies on a low-dimensional set of measurements as inputs. Hence the increased one-time cost to train with a differentiable simulator compared, e.g., to fully supervised training, can pay off significantly over the course of a large number of evaluations when the controller is deployed.

\section{Conclusions}

We have studied the use of differentiable simulations to train neural networks acting as controllers for  complex dynamical systems. The considered system describes the movement of a rigid body subjected to nonlinear perturbations derived from the Navier Stokes equations, posing a very challenging control task.
The proposed approach, which introduced a set of physically interpretable loss terms, is able to train robust controllers without the need to provide reference data for training while relying on a sparse set of measurements. 
It is able to produce controllers that generalizes very well to substantially different, challenging flow conditions.
Numerous interesting venues for future research exist based on our results, such as exploring the transfer of synthetically trained controllers to real-world environments and using the proposed strategy to develop controllers for other control tasks using differentiable simulators.

\bibliographystyle{abbrvnat}
\bibliography{references}

\section*{Checklist}

\appendix

\newcommand{\figsizes}{0.7}
\newcommand{\figsizeshalf}{0.25}

\section{Linear Controllers}


\subsection{PID}
The P gain was adjusted so that the maximum control effort from the PID controller is in the same order of magnitude as the one from the network controllers. The D gain is then tuned to be as low as possible while still avoiding overshoot and the same is done for the I gain. The gains for the controllers for each setup are displayed in Table~\ref{tab:pid_gains} and the control effort $U$ is obtained by 

\begin{align}
    U^t = P e^t + D \frac{e^t - e^{i-1}}{0.1} + 0.1 \sum_{n=0}^{i} e^n  I
\end{align}

\begin{table}[ht]
\caption{Gains of PID controller.}
\label{tab:pid_gains}
\begin{center}
\begin{tabular}{cccc}
\toprule
  & \textbf{P} & \textbf{D} & \textbf{I}  
  \\ \hline \\
Cylinder - Forces  & 1 & 8 & 0.001 \\
Box - Forces  & 2 & 15 & 0.001 \\
Box - Torque  & 100 & 1000  & 0.01 \\
\bottomrule
\end{tabular}
\end{center}
\end{table}

\subsection{Loop Shaping}
Loop Shaping design  consists of finding a controller $K$ so that the open loop response $K\ *\ P$, where $P$ is the system plant, behaves as close as possible to the open loop response of a transfer function $P'$, which is chosen by the user. 

A common choice for $P'$ is $P'(s) = \frac{\omega_b }{s}$ where $\omega_b$ is the control bandwidth and $s$ is a complex frequency. This function has the property of having high gains for low frequencies and low gains for high frequencies. In other words, an input signal with frequency higher than $w_b$ (noise) will be dampened and an input signal with frequency lower than $w_b$ (perturbations) will be amplified. 


We use a combination of two loop-shaping methods: mixed-sensitivity-design, which favors performance, and the Glover-McFarlane method, which favors robustness to plant uncertainty, as implemented by the Matlab \textit{loopsyn()} function. 
It features a parameter $\alpha$: For $\alpha=0$ the controller has the best performance and while $\alpha=1$ favors robustness. 
%
From a range of experiments with our physical environment, we choose $\alpha=0.95$ and $w_b=0.2$. After converting the found controller from the Laplace domain to the discrete one, we obtain the coefficients shown in Table~\ref{tab:ls_coeffs} and the control effort $U$ is calculated according to

\begin{align}
    U^t = \sum_{p=0}^{2} e^{i-p} n_{p} - \sum_{p=1}^{2} U^{i-p} d_{p}
\end{align}
where $i$ is the index of the current time step, $e$ is the error and $n$ and $d$ are the controller coefficients.

\begin{table}[!ht]
\caption{Loop shaping coefficients.}
\label{tab:ls_coeffs}
\begin{center}
\begin{tabular}{ccc}
\toprule
 $p$ & $n_{p}$ & $d_{p}$ 
 \\ \hline \\
0  &  $1.1700924033918623\mathrm{e}{00}$ & - \\
1 & $-1.4694211940919182\mathrm{e}{00}$ & $-1.2306775904257603\mathrm{e}{00}$\\
2 & $3.0598060140064326\mathrm{e}{-01}$ & $2.6726488821832250\mathrm{e}{-01}$\\
\bottomrule
\end{tabular}
\end{center}
\end{table}

\section{Networks Architecture Details}

\begin{figure}[!ht]
    \centering
    \includegraphics[width=\figsizes\linewidth]{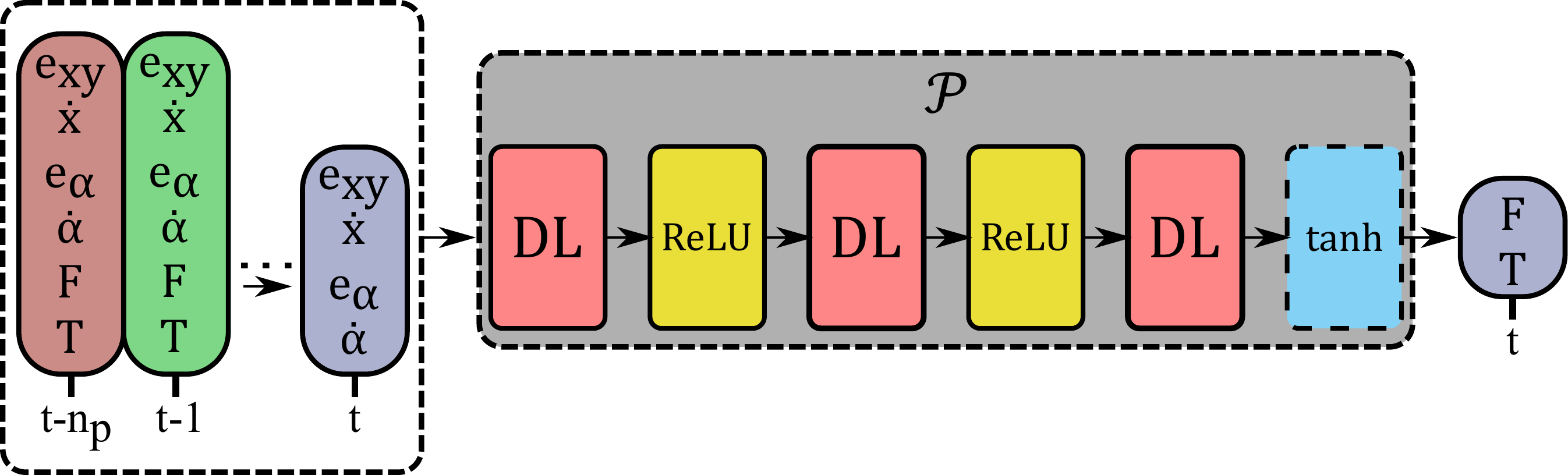}
    \caption{Visualization of the neural network architecture.}
    \label{fig:architecture}
\end{figure}

The dimensions of the considered neural network layers are given in Table~\ref{tab:network}, and
its architecture is visualized in Figure \ref{fig:architecture}. 
The network of the 2 DOF setup has a total of 2206 trainable parameters while the one used in the 3 DOF setup has 2243 trainable parameters. 

\begin{table}[!ht]
\caption{Input/output sizes of neural network layers.}
\label{tab:network}
\begin{center}
\begin{tabular}{ccc}
\toprule
 \textbf{Layer} & \textbf{2 DOF} & \textbf{3 DOF}  
 \\ \hline \\
0 & [16, 38] & [32, 32] \\
1 & [38, 38]  & [32, 32] \\
2 & [38, 2]   & [32, 3]   \\
\bottomrule
\end{tabular}

\end{center}
\end{table}

\section{Training Details}
\subsection{Differentiable Physics} \label{app:diffphys}
The network training parameters for all setups investigated with differentiable physics can be found in Table~\ref{tab:online_hp}. A learning rate of 0.01 was used in our tests. Additionally, learning rate decay is used so that the learning rate drops to half of its initial value every 200 and 1000 training iterations for the 2 DOF and 3 DOF setups, respectively.
The networks inputs and outputs are normalized based on a set of measured simulation statistics.

\begin{table}[!ht]
\caption{Parameters of network trained with differentiable physics.}
\label{tab:online_hp}
\begin{center}
\begin{tabular}{ccc}
\toprule
 \textbf{Hyperparameter} & \textbf{2 DOF} & \textbf{3 DOF}  
 \\ \hline \\
$\beta_{xy}$ & 15 & 5 \\
$\beta_{\dot{x}}$ & 5  & 5  \\
$\beta_{F}$ & 0.1   & 0.1   \\
$\beta_{\Delta F}$ & 2 & 1 \\ 
$\beta_{prox}$ & 0.1 & 0.1\\ 
$\beta_{\alpha}$ & - & 30\\ 
$\beta_{\dot{\alpha}}$ & - & 0.05\\ 
$\beta_{\Delta T}$ & - &  1 \\
$n_{i}$ & 1000 & 5000 \\ 
$l$ & 16 & 16 \\ 
\bottomrule
\end{tabular}

\end{center}
\end{table}



\subsubsection{Effect of Time Horizon l}
\label{sec:time_horizon}
\begin{figure}[!ht]
\centering
    \begin{subfigure}{\figsizes\linewidth}
        \centerline{\includegraphics[width=\linewidth]{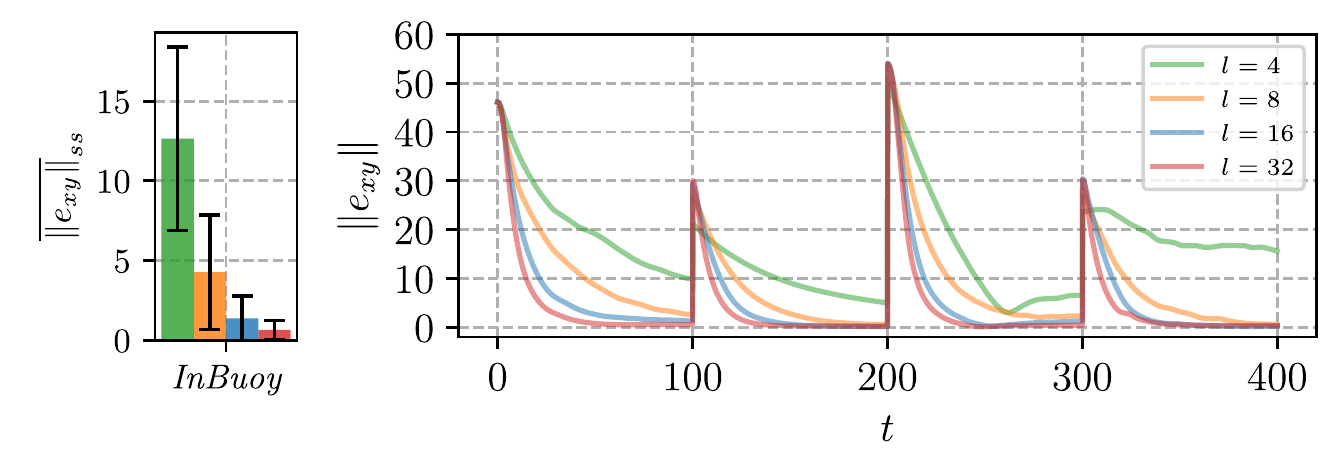}}
    \end{subfigure}
    
    \begin{subfigure}{\figsizes\linewidth}
        \centerline{\includegraphics[width=\linewidth]{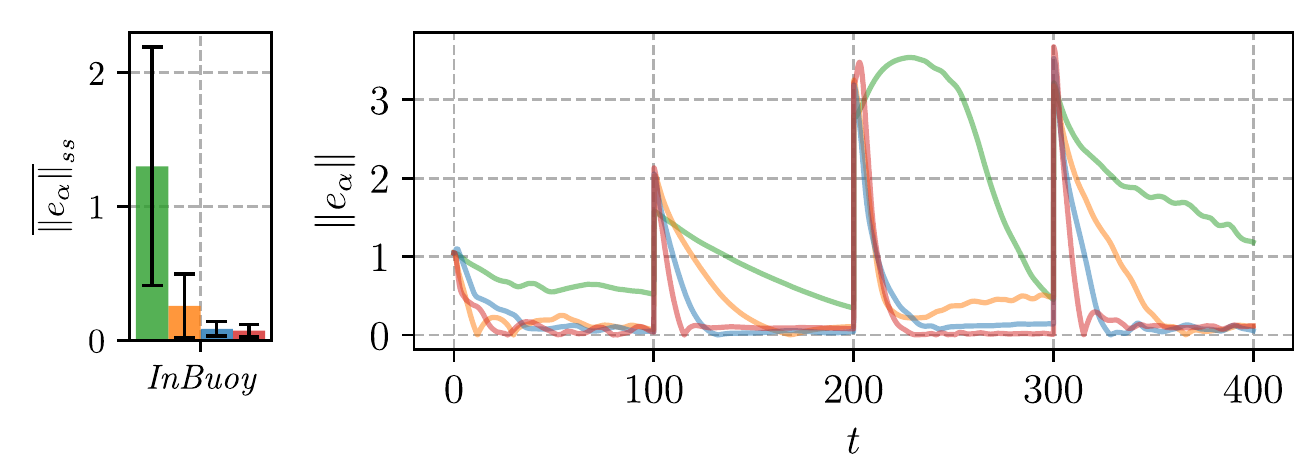}}
    \end{subfigure}
    
    \caption{Steady state errors (left) and errors norms (right) for a test with \inflowBuoyancyThree parameters. Training with small time horizons $l$ produces poor performant controllers. When using a very large time horizon, such as $l=32$, the gains are not worth the increased training times.}
    \label{fig:unrolling_study}
\end{figure}

The effect of training with different time horizons $l$ is also investigated. A test consisting of one simulation with parameters \inflowBuoyancyThree and a \textit{N} shaped trajectory is conducted and the performance of networks trained with different values of $l$ can be seen in Figure \ref{fig:unrolling_study}. Small values of $l$ mean that loss and backpropagation are performed for short physical timespans during training, i.e. a small temporal lookahead, which produces a controller with a deteriorated performance. 
In addition, performance does not change significantly beyond $l=16$. Since the same number of iterations were conducted for all runs, using larger time horizons only increases training time. Therefore, we choose $l=16$ since it provides a good balance between performance and training time. 

\subsubsection{Sensitivity to Weights Initialization}
\label{sec:seeds}
\begin{figure}[!ht]
    \centering
    \begin{subfigure}{\figsizes\linewidth}
        \centerline{\includegraphics[width=\linewidth]{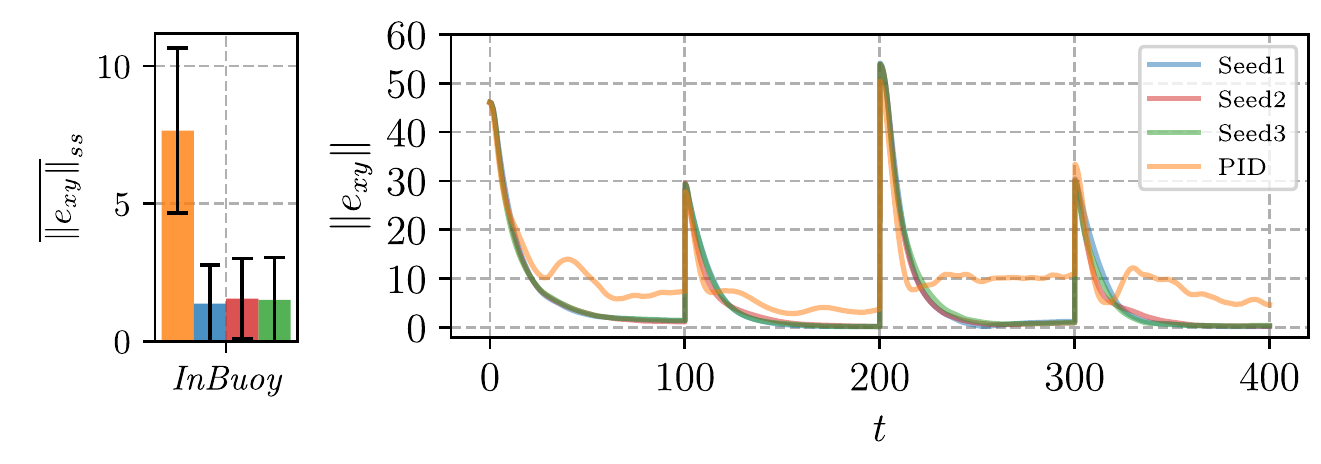}}
    \end{subfigure}
    
    \begin{subfigure}{\figsizes\linewidth}
        \centerline{\includegraphics[width=\linewidth]{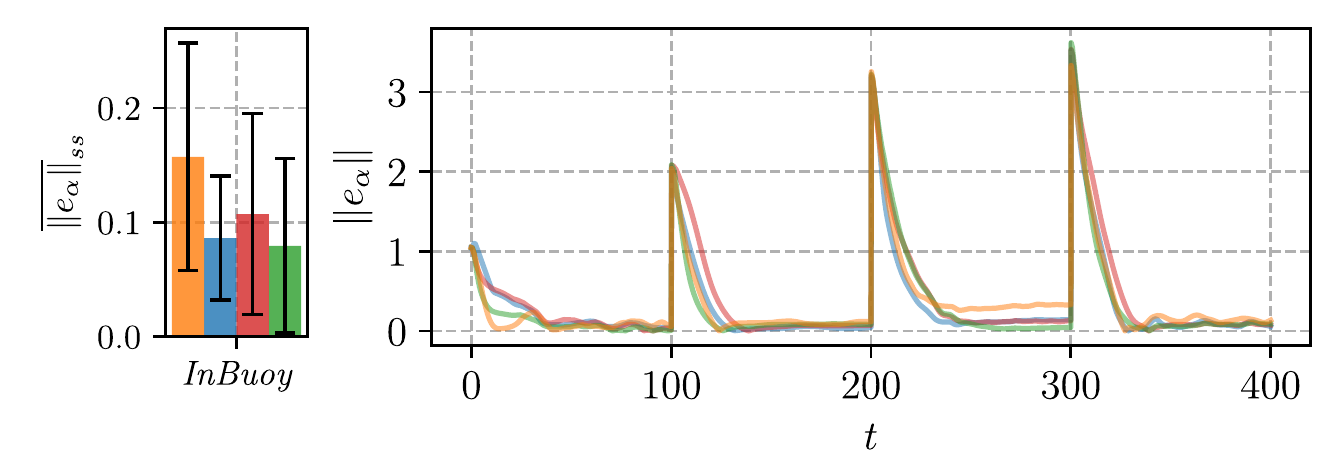}}
    \end{subfigure}
    
    \caption{Steady state errors (left) and errors norms (right) for a test with \inflowBuoyancyThree parameters. The networks have the same parameters but different initialization seeds. Similar tracking performance is achieved for all of them, with Seed 2 having a slightly worse angular tracking.}
    \label{fig:seeds}
\end{figure}

In order to evaluate how the performance from the network trained with differentiable physics is influenced by the networks weights initial values, we perform three training runs with different initial seeds. We run a test simulation with parameters from \inflowBuoyancyThree and objectives that describe a \textit{N} shape trajectory. Despite minor differences the error norms for all seeds have a similar tracking performance. This is especially apparent when comparing it to other controllers, as shown in Figure \ref{fig:seeds}.

\subsubsection{Ablation Studies}
\label{sec:ablation}
\begin{figure}[!ht]
    \centering
    \begin{subfigure}{\figsizes\linewidth}
        \centerline{\includegraphics[width=\linewidth]{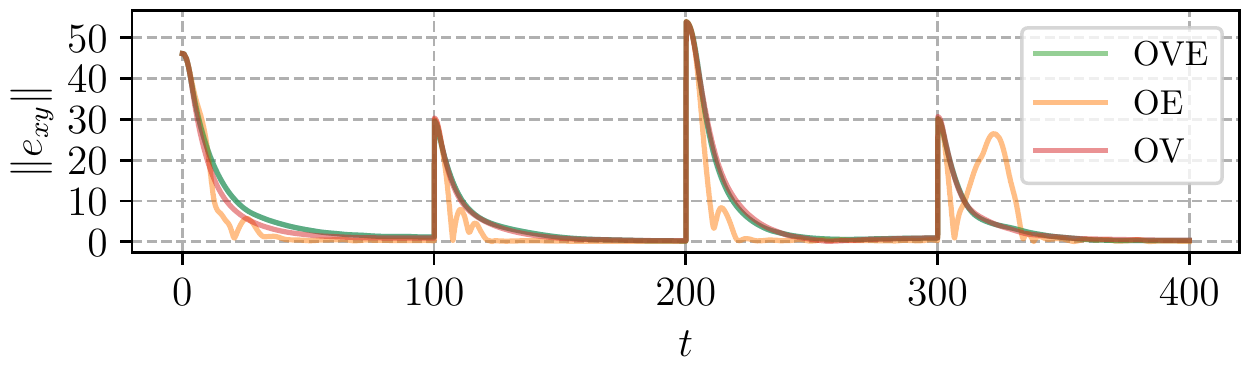}}
    \end{subfigure}
    
    
    \begin{subfigure}{\figsizes\linewidth}
        \centerline{\includegraphics[ width=\linewidth]{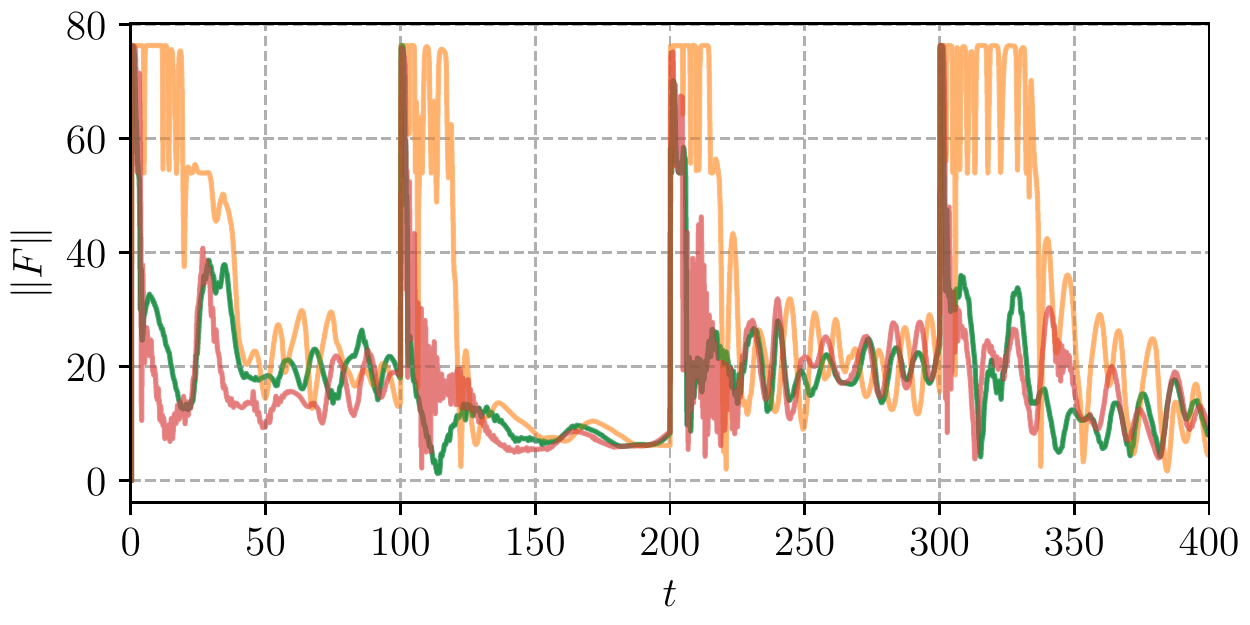}}
    \end{subfigure}  
    
    \begin{subfigure}{\figsizes\linewidth}
        \centerline{\includegraphics[ width=\linewidth]{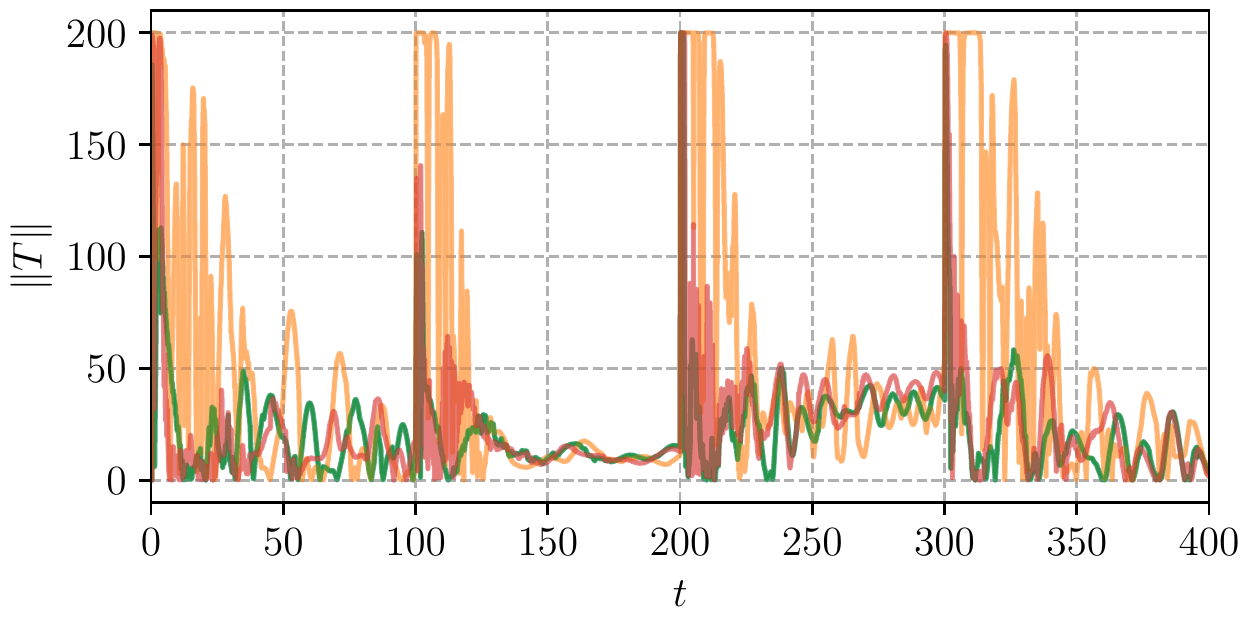}}
    \end{subfigure}   
    
    \begin{subfigure}{\figsizes\linewidth}
        \centerline{\includegraphics[width=\linewidth]{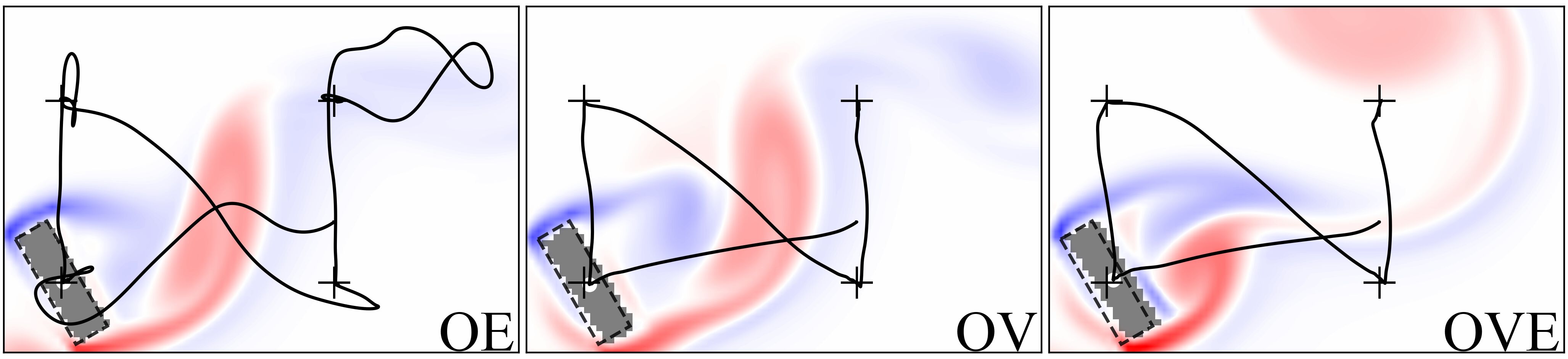}}
    \end{subfigure}

    \caption{From top to bottom: errors norm, control efforts norm and trajectories of controllers colored by vorticity at $t=95$ on test with  \inflowBuoyancyThree parameters (bottom). All loss terms are necessary in order to accomplish steady error decay and maintain small and smooth control efforts.}
    \label{fig:ablation}
\end{figure}

We present an evaluation study with simulations performed with \inflowBuoyancyThree parameters in order to assess the influence of the terms introduced in Section \ref{sec:ol}. One simulation  is performed for each combination of loss terms.
When using only the \textit{O} term, the training is severely under constrained, and as a consequence the network learns to exert overly large forces.
This destabilizes the simulations preventing a successful training. 
Therefore we compare the performance of networks trained with the combinations: \textit{OVE} (all terms), \textit{OV} (objective and velocity terms) and \textit{OE} (objective and effort terms). 
Networks trained with \textit{OVE} and \textit{OV} exhibit similar error curves and trajectories as shown in Figure \ref{fig:ablation}. The network trained with \textit{OE} is also able to achieve low error values but it exhibits an uneven decay, which is an undesirable behavior. When analyzing the control efforts it can be seen that \textit{OE} provides a network that produces maximum control efforts more often, which means it uses more energy. When considering \textit{OV}, there are no constraints on control efforts and consequently it learns a control policy that modulates the control efforts with a high frequency - an undesirable behavior in practice. 


\subsection{Supervised Learning} \label{app:super}
\begin{figure}[!ht]
    \centering
    \begin{subfigure}{\figsizes\linewidth}
        \centerline{\includegraphics[width=\linewidth]{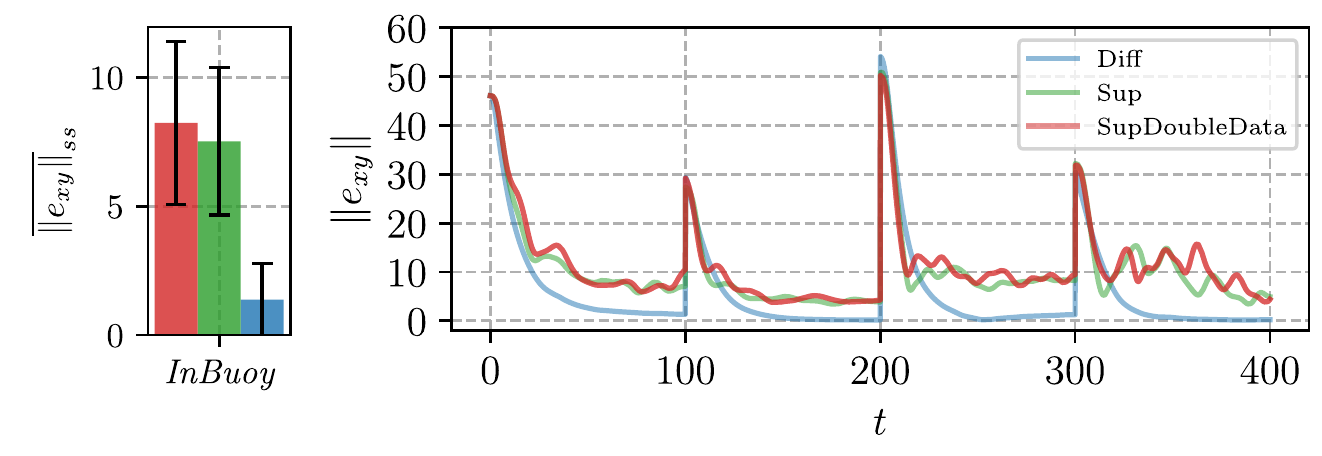}}
    \end{subfigure}
    
    \begin{subfigure}{\figsizes\linewidth}
        \centerline{\includegraphics[width=\linewidth]{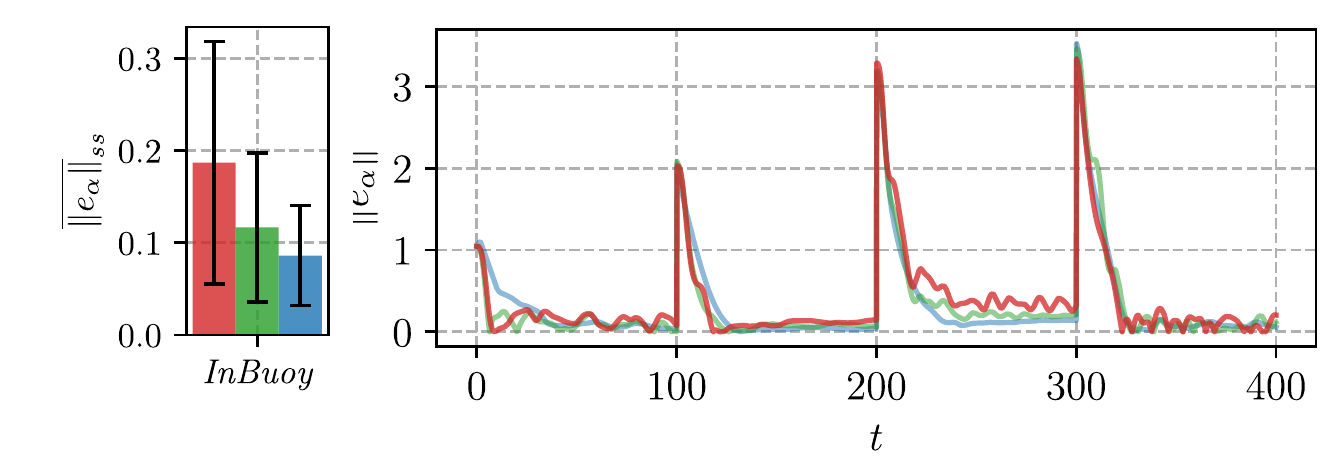}}
    \end{subfigure}
    
    \caption{Steady state errors (left) and errors norms (right)
for a test with \inflowBuoyancyThree parameters. A larger dataset does not improve model performance.}
    \label{fig:supervised}
\end{figure}

In total $n_i=150{,}000$ iterations are performed. A learning rate of 0.01 is used with learning rate decay so that every $15{,}000$ iterations it drops to half of its size. 
Training with more data is examined for assessing if the performance of the controller acquired via supervised learning could be improved. A dataset with 200 simulations (double the size of the original one) is generated with 180 simulations being used for training and 20 for validation. Then a test simulation with parameters from \inflowBuoyancyThree and objectives describing a \textit{N} shaped trajectory is conducted. It can be seen that adding more data did not result in clear improvements, as shown in Figure \ref{fig:supervised}.

\subsection{Reinforcement Learning} \label{app:rl}

We use the SAC implementation from stable-baselines3 (\citet{stable-baselines3}). 
A fixed learning rate of 0.0003 and a reward discount factor $\gamma$ of 0.99 were chosen. The parameter $\tau$ controlling the Polyak Averaging of the two Q-Functions within the critic is set to 0.05. The training takes place until $n_i=500{,}000$ iterations are performed.

\subsection{Hardware and Software}
\label{sec:hardware}
All optimization procedures were conducted utilizing the PyTorch framework (\citet{Pytorch}) on a GeForce RTX 2080 Ti. Approximate training times are displayed in Table~\ref{tab:training_times}.

\begin{table}[!ht]
\caption{Training times.}
\label{tab:training_times}
\begin{center}
\begin{tabular}{ccc}
\toprule
\textbf{Algorithm} & \textbf{2 DOF} & \textbf{3 DOF} 
\\ \hline \\
Reinforcement & 21h & - \\
Supervised & 0.5h & 1h  \\
Diff. Physics & 0.6h & 4h \\ 
\bottomrule
\end{tabular}
\end{center}
\end{table}

\section{Additional Results}\label{app:results}
In the following we present a collection of additional results that were not shown in the main body of the paper. For all simulations, the cylinder has $m=11.78$ and radius $\hat{r}=5$ while the box has $m=36$, $I=4000$, width $w=20$ and height $h=6$.



\begin{figure}[!ht]
    \centering
    \begin{subfigure}{\figsizeshalf\linewidth}
        \centerline{\includegraphics[width=\linewidth]{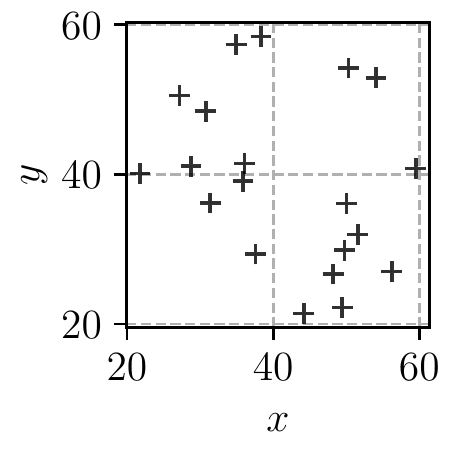}}
    \end{subfigure}
    \begin{subfigure}{\figsizeshalf\linewidth}
        \centerline{\includegraphics[width=\linewidth]{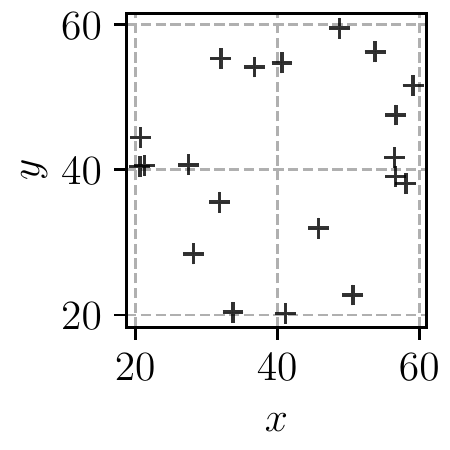}}
    \end{subfigure}
    
    \begin{subfigure}{\figsizes\linewidth}
        \centerline{\includegraphics[width=\linewidth]{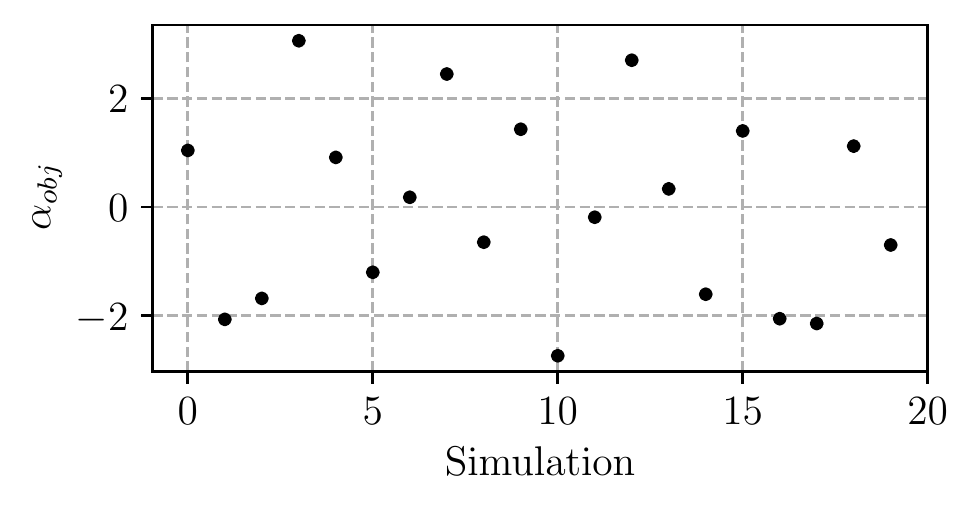}}
    \end{subfigure}
    
    \caption{Targets of 2 DOF validation (top-left) and 3 DOF validation (top-right) tests as well as angle targets of 3 DOF validation test (bottom) . The rigid body initial position is located at $(x,y)=(40,40)$.}
    \label{fig:objectives}
\end{figure}

\begin{figure}[!ht]
\centering
    \begin{subfigure}{0.7\linewidth}
        
        \centerline{\includegraphics[width=\linewidth]{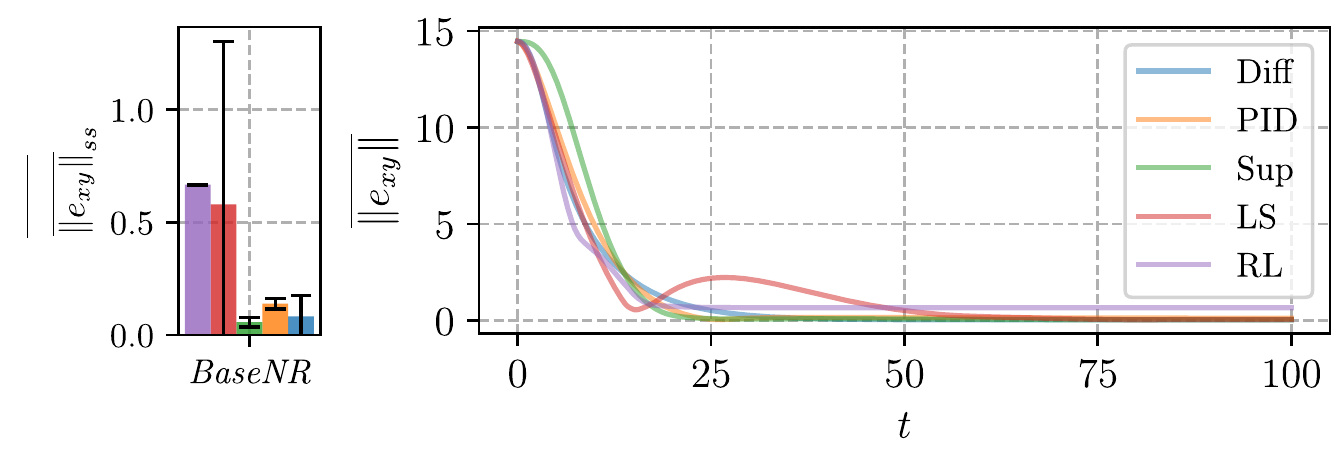}}

    \end{subfigure}
    %
    
    \begin{subfigure}{0.7\linewidth}
        \centering
    \includegraphics[width=\linewidth]{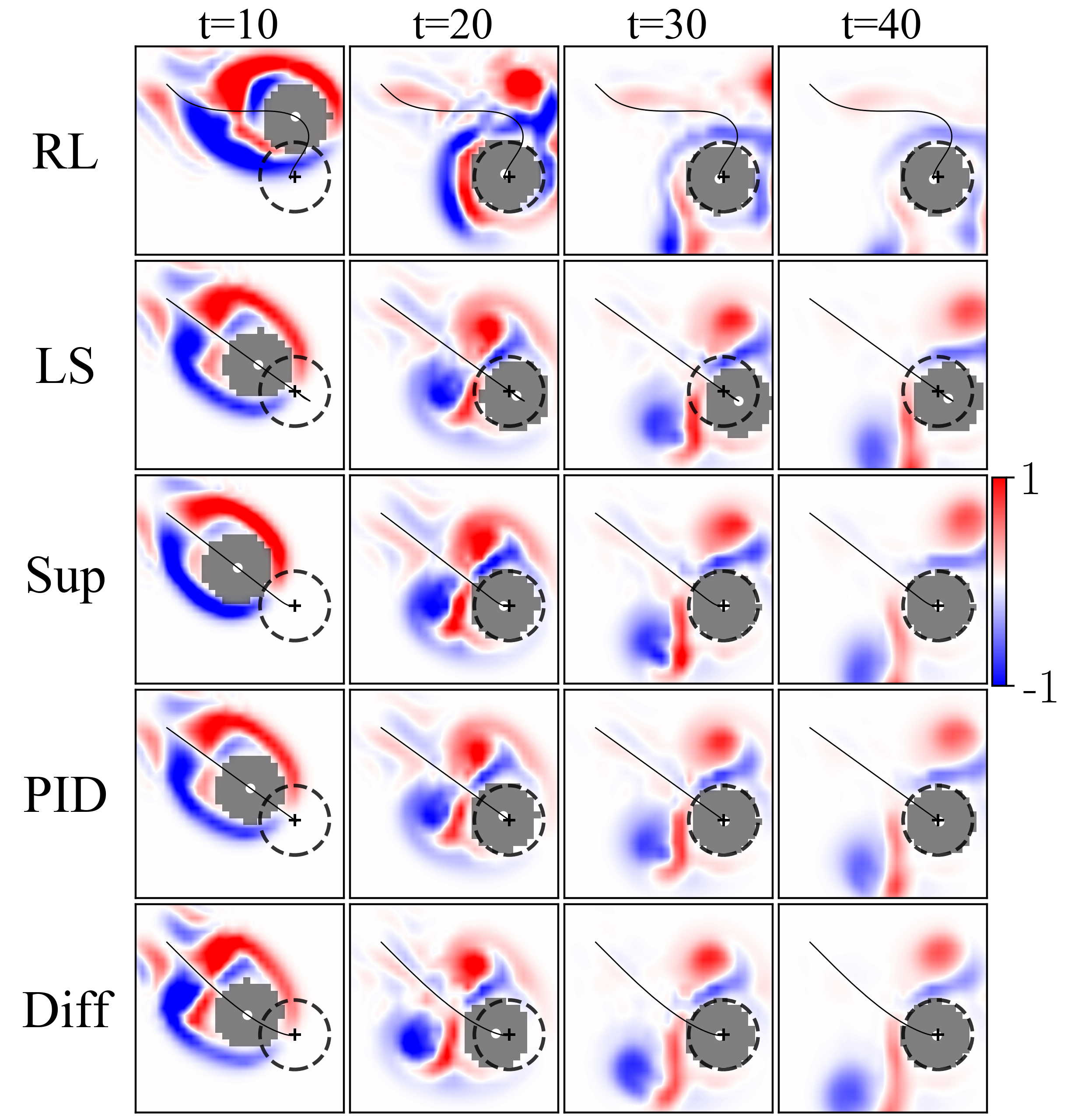}
    \end{subfigure}
    \caption{Average spatial error norm of validation simulations with \baselineTwo parameters (top-left) and average steady state error (top-right). All controllers achieve spatial steady state errors smaller than one. Vorticity contours of one simulation from 2 DOF validation test (bottom).}
    \label{fig:ba2}
\end{figure}

\begin{figure}[!ht]
    \centering
    \begin{subfigure}{0.2\linewidth}
        \centerline{\includegraphics[width=1.1\linewidth]{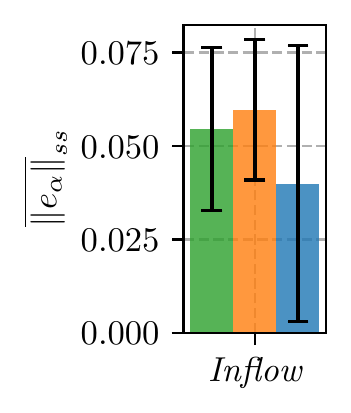}}
    \end{subfigure}
    \begin{subfigure}{0.2\linewidth}
        \centerline{\includegraphics[width=\linewidth]{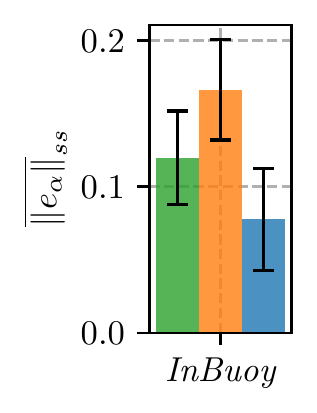}}
    \end{subfigure}
    \begin{subfigure}{0.2\linewidth}
        \centerline{\includegraphics[width=\linewidth]{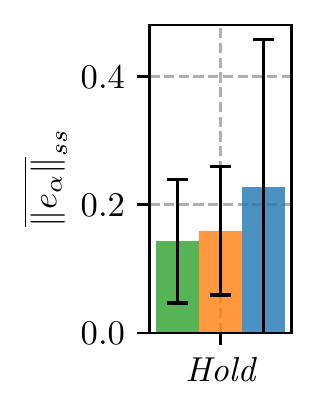}}
    \end{subfigure}
    \caption{Average angular steady state errors of tests performed with \inflowThree, \inflowBuoyancyThree and \hold parameters (from left to right, respectively).}
\end{figure}

\begin{table*}[!ht]
\caption{Spatial steady state error and standard deviation of the considered tests.}
\label{tab:spatial_ss}
\begin{center}
\vskip -0.1in
\resizebox{\linewidth}{!}{
\begin{tabular}{cccccc}
\toprule
\textbf{Test} & \textbf{RL} & \textbf{LS} & \textbf{Sup} & \textbf{PID} & \textbf{Diff}  
\\ \hline \\
Validation 2 DOF & $0.6673 \pm  \mathbf{0.0007}$ & $0.5797 \pm 0.7233$ & $\mathbf{0.0563} \pm 0.0203$ & $0.1386 \pm 0.0227$ & $0.08247 \pm 0.0924$  \\
Validation 3 DOF & - & - & $\mathbf{0.2115} \pm 0.2138$ & $0.2206 \pm \mathbf{0.2124}$ & $0.7762 \pm 0.6004$ \\
Test with \buoyancyTwo & $0.7634 \pm 0.2587$ & $9.6079 \pm 2.1461$ & $2.9695 \pm 0.6410$ & $2.2122 \pm 0.4086$ & $\mathbf{0.1943} \pm \mathbf{0.1800}$ \\ 
Test with \inflowThree & - & - & $3.6632 \pm \mathbf{0.6446}$ & $3.7965 \pm 0.7141$ & $\mathbf{1.4081} \pm 1.3670$ \\ 
Test with \inflowBuoyancyThree & - & - & $7.1705 \pm \mathbf{0.9169}$ & $8.0952 \pm 1.2625$ & $\mathbf{1.5611} \pm 1.3261$ \\
Test with \hold & - & - & $16.2431 \pm 7.1467$ & $15.0201 \pm 6.2401$ & $\mathbf{4.4825} \pm \mathbf{5.7908}$ \\ 
\bottomrule
\end{tabular}
}
\end{center}
\end{table*}

\begin{table*}[!ht]
\caption{Angular steady state error and standard deviation of tests.}
\label{tab:angle_ss}
\begin{center}
\begin{tabular}{cccc}
\toprule
\textbf{Test} & \textbf{Sup} & \textbf{PID} & \textbf{Diff}  
\\ \hline \\
Validation 3 DOF & $\mathbf{0.0142} \pm \mathbf{0.0218}$ & $0.0163 \pm 0.0271$ & $0.0183 \pm 0.0239$ \\ 
Test with \inflowThree & $0.0545 \pm 0.0218 $ & $0.0596 \pm \mathbf{0.0187}$ & $\mathbf{0.0399} \pm 0.0369$ \\
Test with \inflowBuoyancyThree & $0.1198 \pm \mathbf{0.0318}$ & $0.1662 \pm 0.0343$ & $\mathbf{0.0775} \pm 0.0348$ \\ 
Test with \hold & $\mathbf{0.1430} \pm \mathbf{0.0962}$ & $ 0.1592 \pm 0.0999$ & $0.2269 \pm 0.2302$ \\ 
\bottomrule
\end{tabular}
\end{center}
\end{table*}

\begin{figure}[!ht]
    \centering
    \begin{subfigure}{\figsizes\linewidth}
        \centerline{\includegraphics[width=\linewidth]{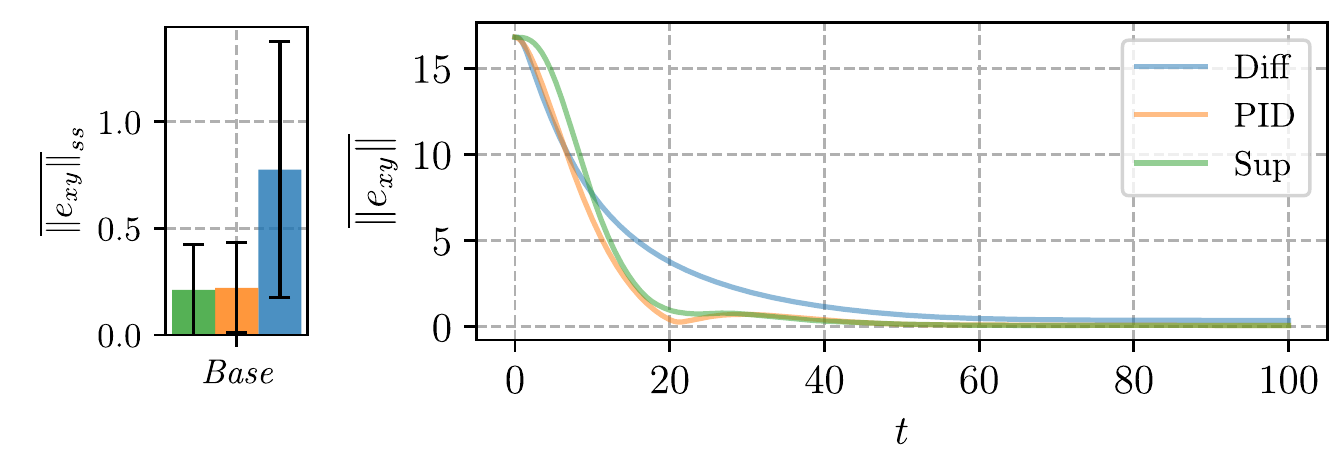}}
    \end{subfigure}

    \begin{subfigure}{\figsizes\linewidth}
        \centerline{\includegraphics[width=\linewidth]{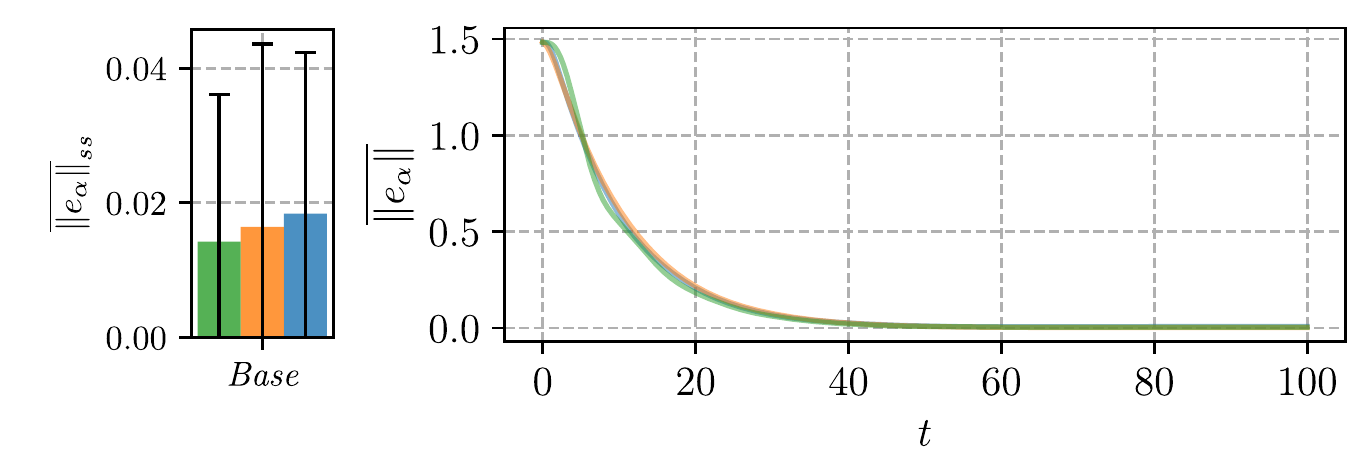}}
    \end{subfigure}
    
    \begin{subfigure}{\figsizes\linewidth}
        \centerline{\includegraphics[width=\linewidth]{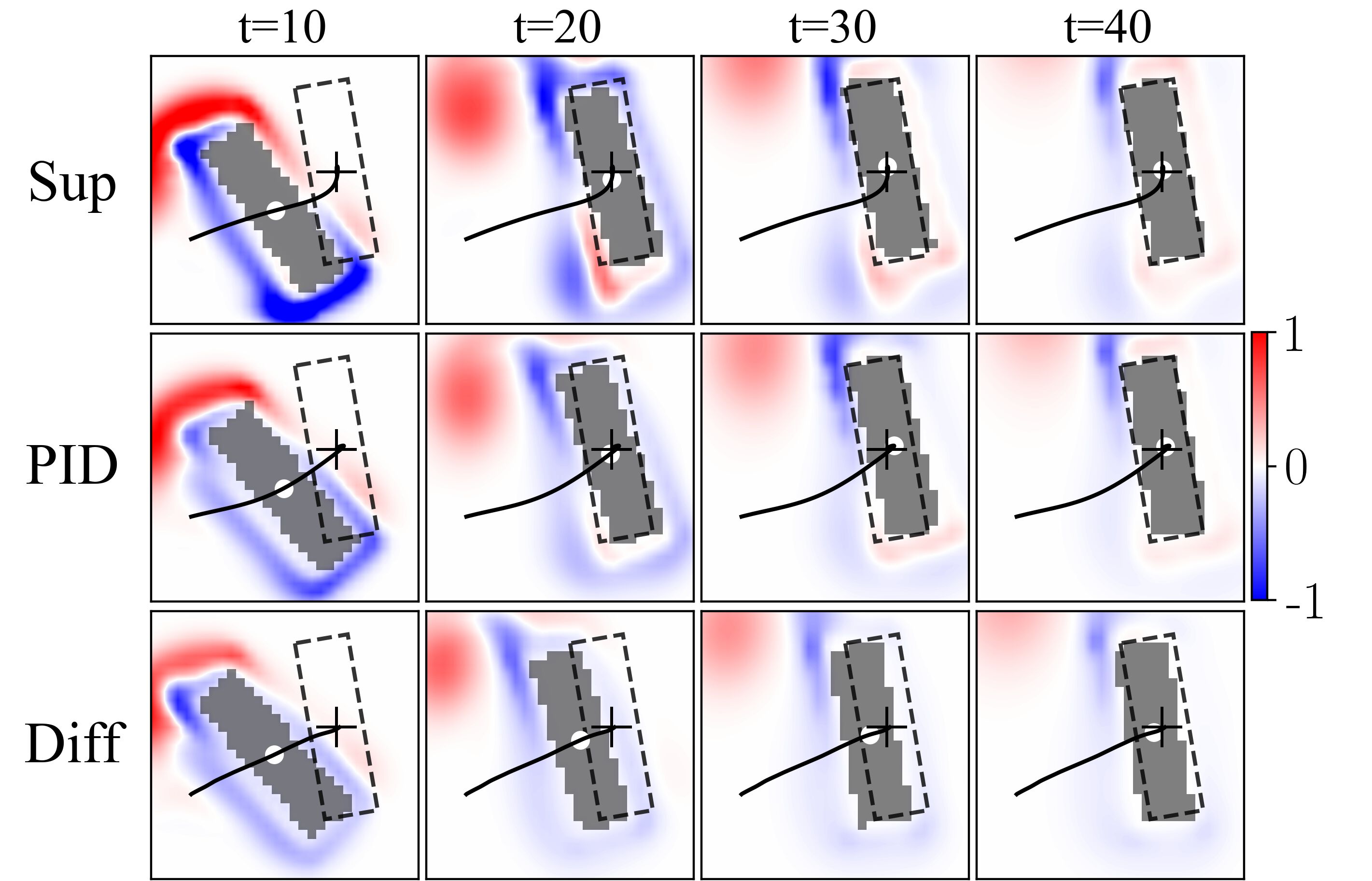}}
    \end{subfigure}
    
    \caption{Average steady state errors (top-left, middle-left),
trajectories of one of the simulations (bottom) and their
error norms (top-right, middle-right) with parameters \baselineThree. All controllers are able to achieve low steady state errors. The increased spatial steady state error from $\mathcal{P}_{\text{diff}}$ is caused by its slower spatial tracking.}
    \label{fig:base3}
\end{figure}


\begin{figure}[!ht]
\centering
\begin{subfigure}[t]{\figsizes\linewidth}
    \centering
    \includegraphics[width=\linewidth]{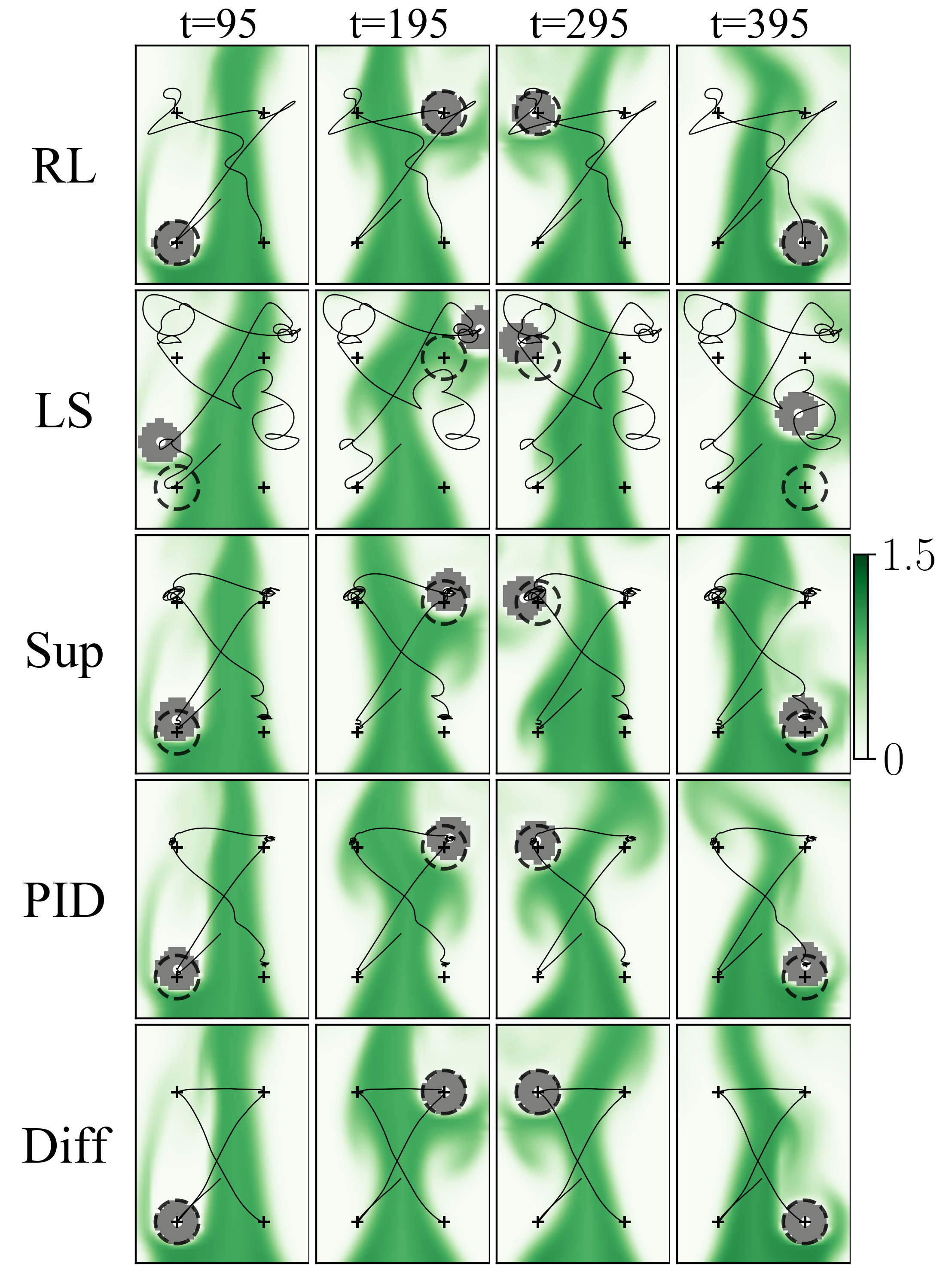}
\end{subfigure}
\begin{subfigure}[t]{\figsizes\linewidth}
    \centering
    \includegraphics[width=\linewidth]{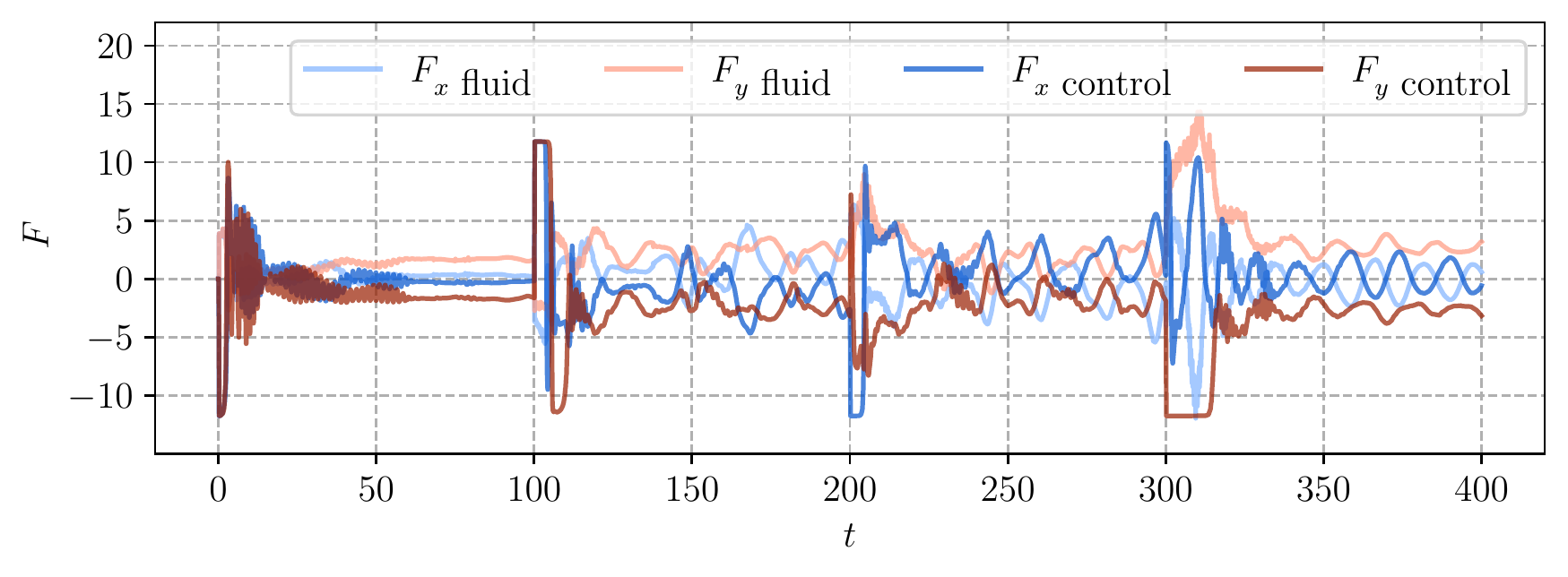}
\end{subfigure}
    \caption{Density contours of lighter fluid (top) of one simulation from test with \buoyancyTwo. Fluid and control forces (bottom) from  $\mathcal{P}_{\text{diff}}$ run. }
\end{figure}

\begin{figure}[!ht]
\centering
\begin{subfigure}[t]{\figsizes\linewidth}
    \centering
    \includegraphics[width=\linewidth]{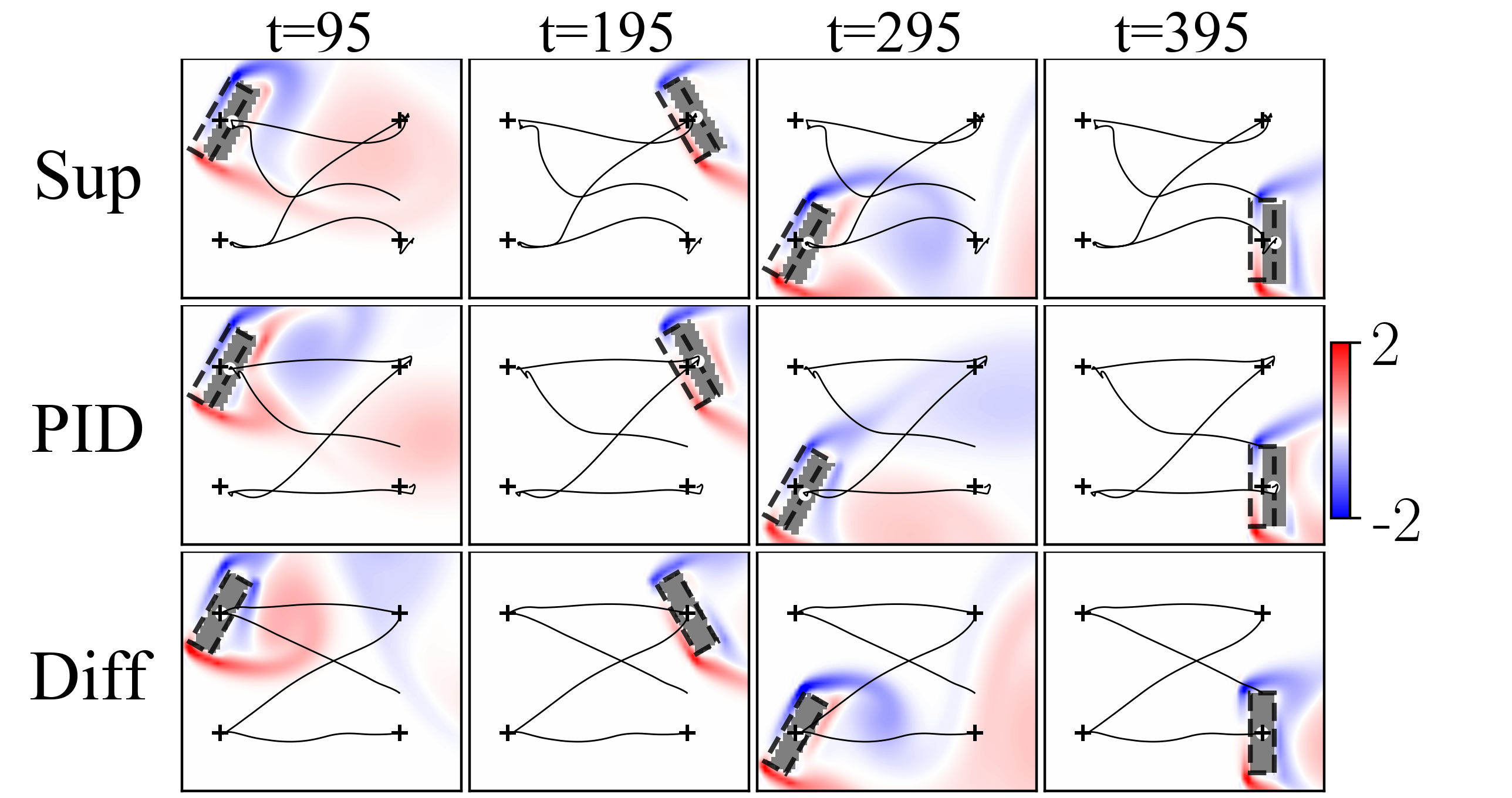}
\end{subfigure}
\begin{subfigure}[t]{\figsizes\linewidth}
    \centering
    \includegraphics[width=\linewidth]{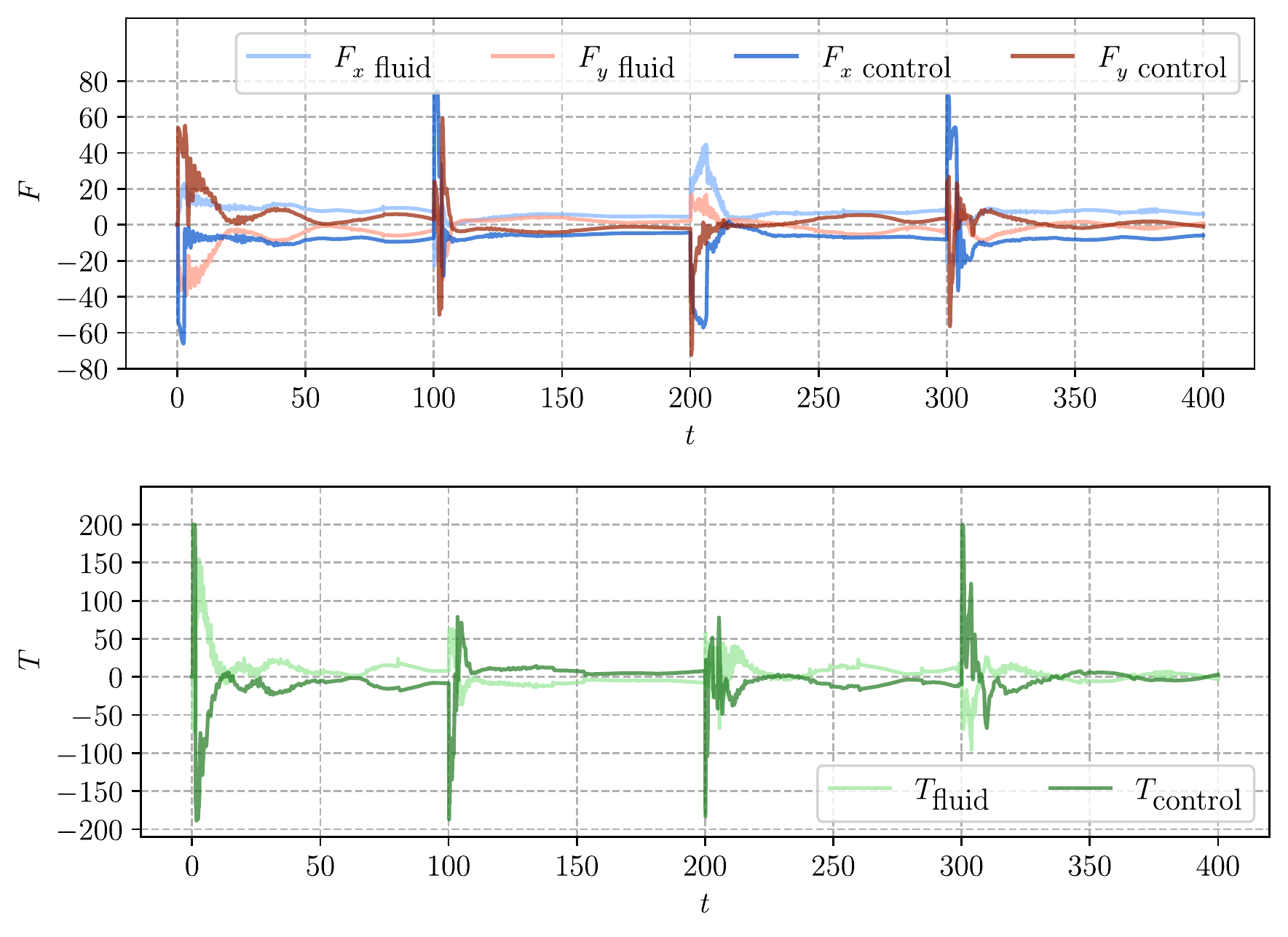}
\end{subfigure}
    \caption{Vorticity contours (top) of one simulation from test with \inflowThree. Fluid and control forces (middle) and torques (bottom) from  $\mathcal{P}_{\text{diff}}$ run. }
\end{figure}

\begin{figure}[!ht]
\centering
\begin{subfigure}[t]{\figsizes\linewidth}
    \centering
    \includegraphics[width=\linewidth]{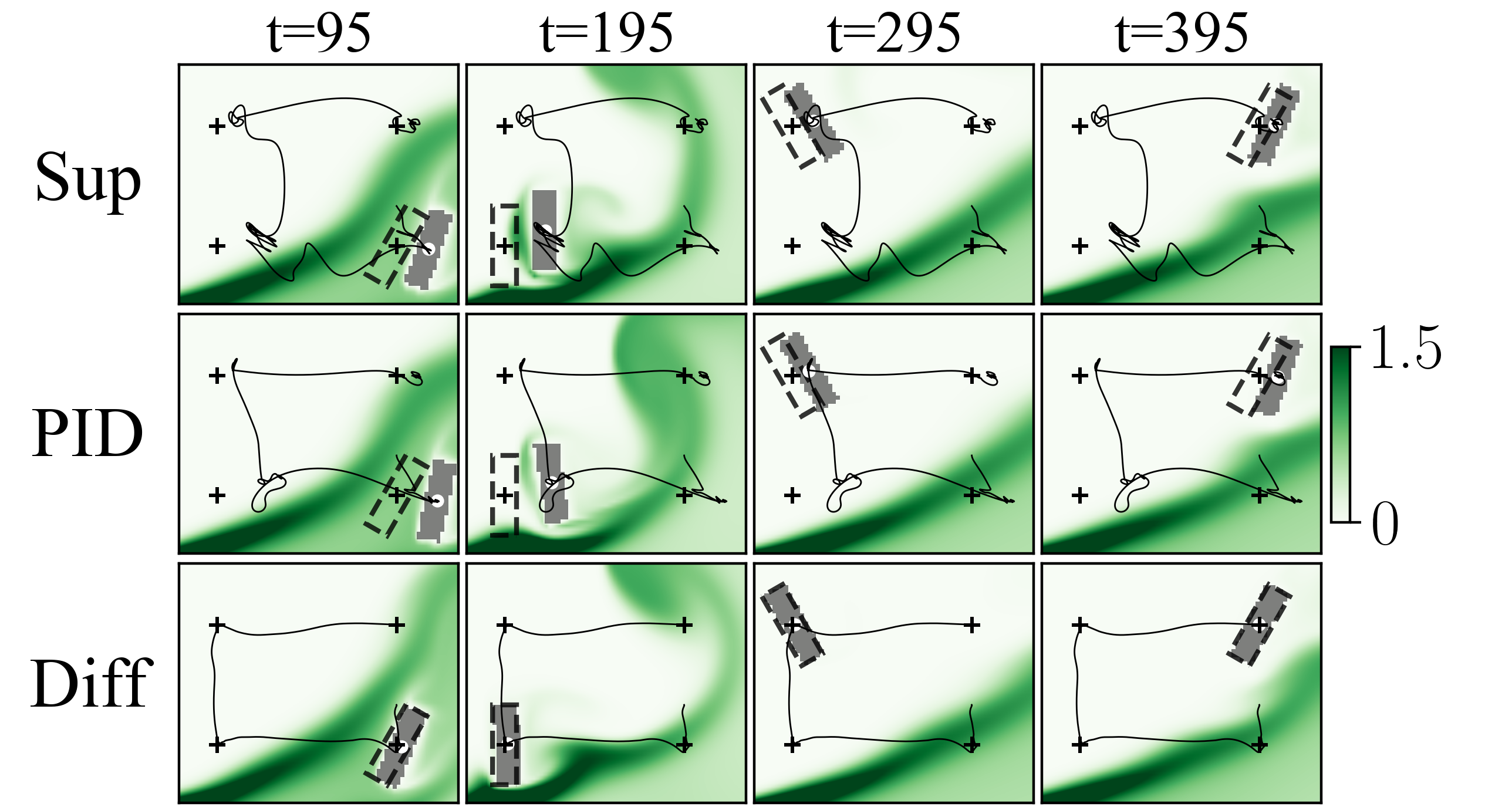}
\end{subfigure}
\begin{subfigure}[t]{\figsizes\linewidth}
    \centering
    \includegraphics[width=\linewidth]{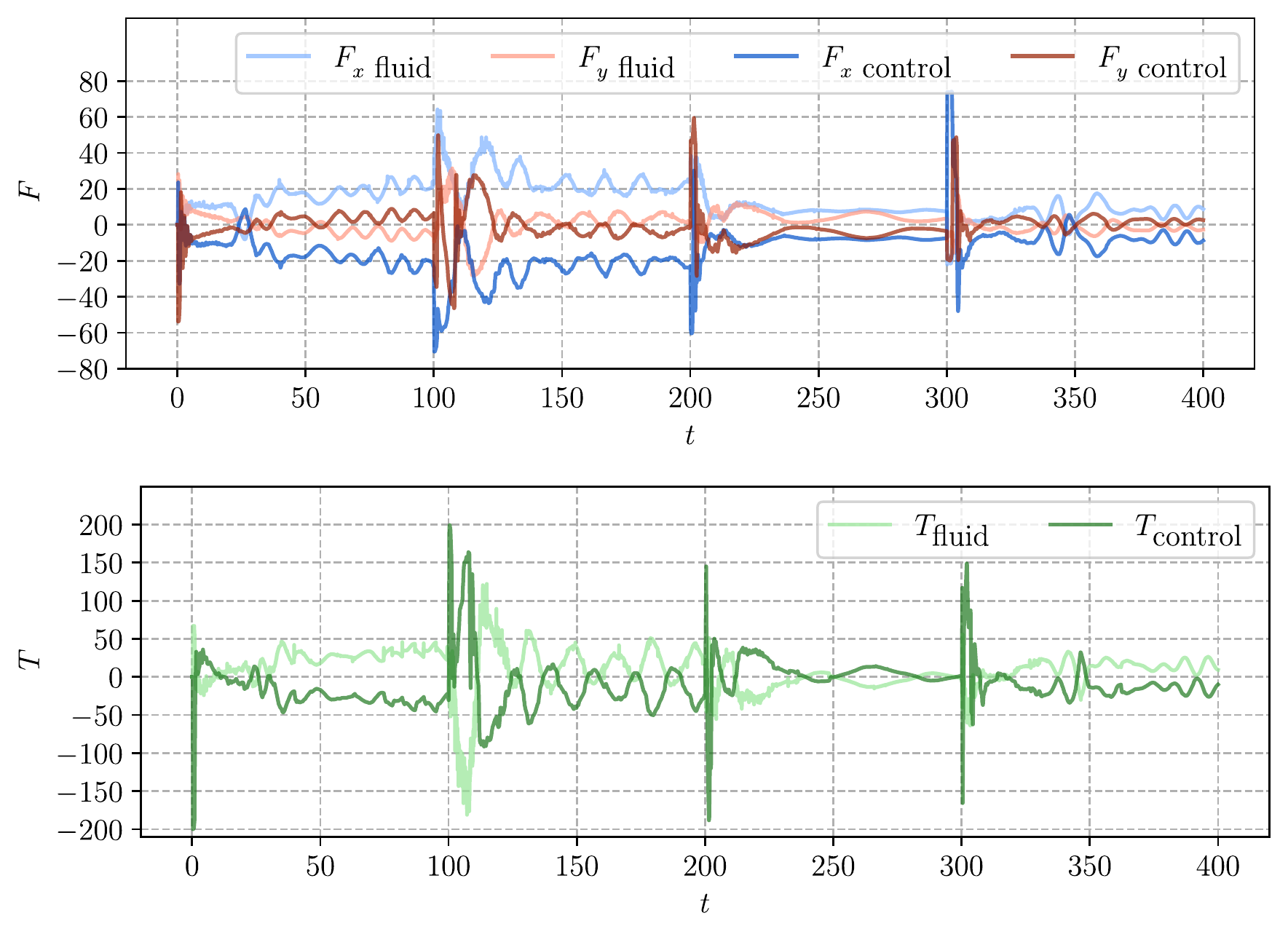}
\end{subfigure}
    \caption{Density contours of lighter fluid (top) of one simulation from test with \inflowBuoyancyThree. Fluid and control forces (middle) and torques (bottom) from  $\mathcal{P}_{\text{diff}}$ run. }
\end{figure}

\begin{figure}[!ht]
\centering
\begin{subfigure}[t]{\figsizes\linewidth}
    \centering
    \includegraphics[width=\linewidth]{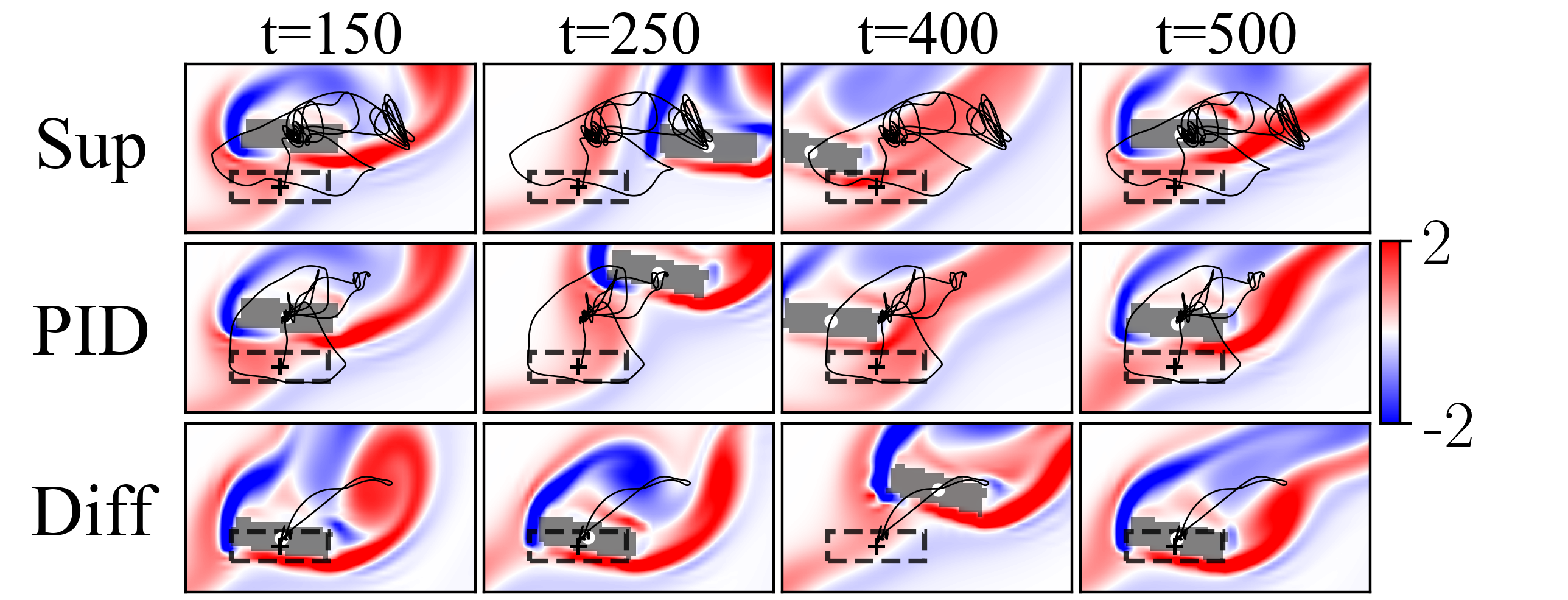}
\end{subfigure}
\begin{subfigure}[t]{\figsizes\linewidth}
    \centering
    \includegraphics[width=\linewidth]{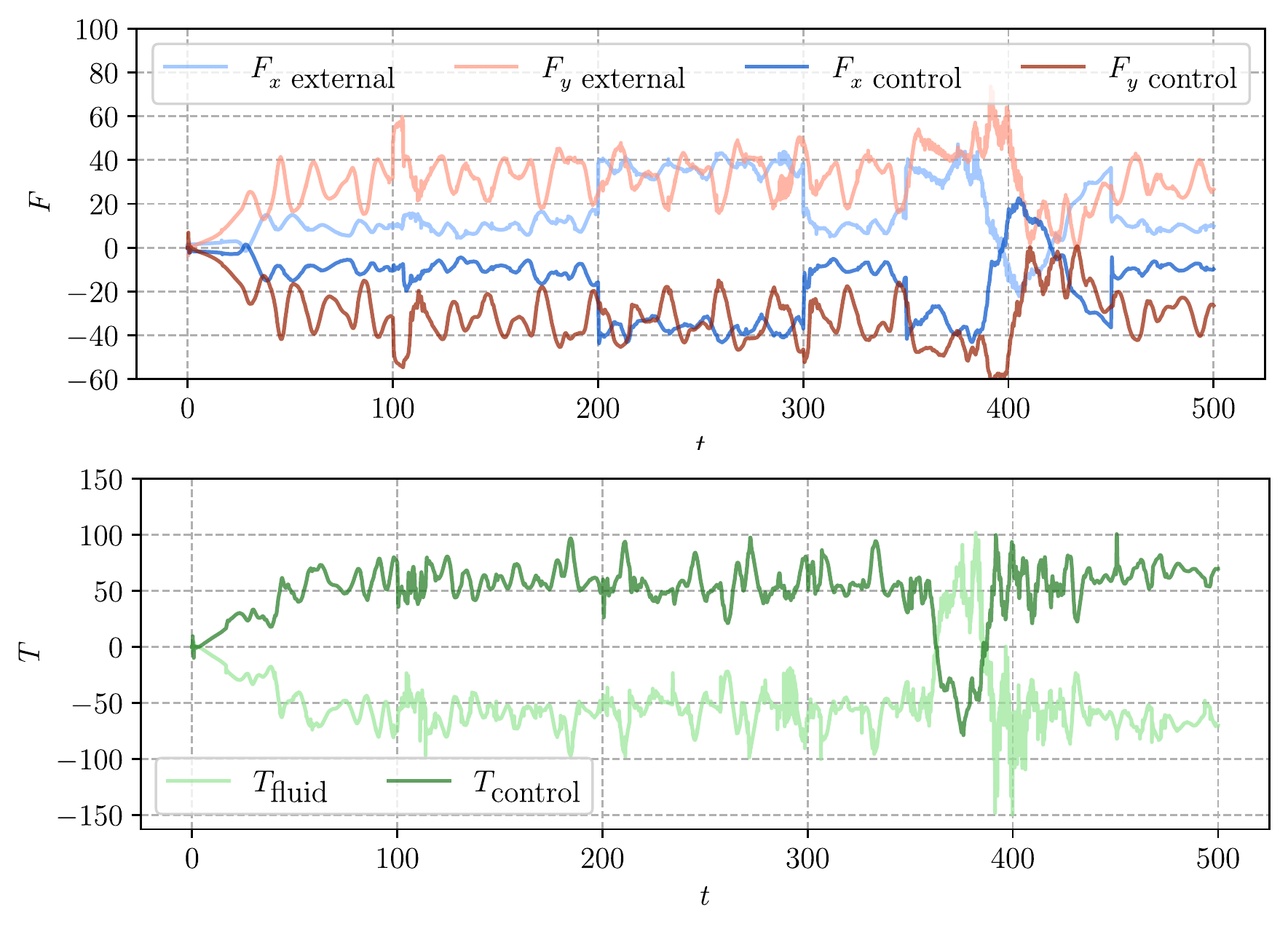}
\end{subfigure}
    \caption{Vorticity contours (top) from test with \hold. External (fluid + forcing) and control forces (middle) and torques (bottom) from  $\mathcal{P}_{\text{diff}}$ run. }
\end{figure}

\end{document}